\documentclass[journal,twocolumn]{IEEEtran}
\usepackage{cite}
\usepackage{amsthm}
\newtheorem{myDef}{Definition}

\newtheorem{lemma}{Lemma}
\usepackage{amsmath,amssymb,amsfonts}
\usepackage{algorithmic}
\usepackage{graphicx}
\usepackage{textcomp}
\usepackage{subfigure}
\usepackage{xcolor}
\usepackage{color}
\usepackage{siunitx}
\usepackage{pifont}
\usepackage{threeparttable}
\usepackage{hyperref}[colorlinks,linkcolor=blue]
\def\BibTeX{{\rm B\kern-.05em{\sc i\kern-.025em b}\kern-.08em
    T\kern-.1667em\lower.7ex\hbox{E}\kern-.125emX}}
\usepackage{balance}

\begin{document}
\title{eRSS-RAMP: A Rule-Adherence Motion Planner Based on Extended Responsibility-Sensitive Safety for Autonomous Driving}

% \author{TBD}% Author list will be determined before the day of submission, which depends on the key contribution of each co-author.

\author{Pengfei Lin, Ehsan Javanmardi,~Yuze~Jiang,~Dou~Hu, Shangkai Zhang,~and~Manabu~Tsukada \vspace{-2em}% <-this % stops a space
\thanks{This work was supported by JST ASPIRE Grant Number JPMJAP2325, Japan, and the Japan Society for the Promotion of Science (JSPS) Research Fellowship for Young Scientists program (grant number: 23KJ0391). (\textit{Corresponding author: Pengfei Lin})}
\thanks{P. Lin, E. Javanmardi, Y. Jiang, D. Hu, S. Zhang, and M. Tsukada are with the Department of Creative Informatics, The University of Tokyo, Tokyo 113-8654, Japan (e-mail: \{linpengfei0609,~ejavanmardi,~uiryuu,~douhu,~zhangshangkai,~mtsukada\}@g.ecc.u-tokyo.ac.jp).}% <-this % stops a space
% <-this % stops a space
}

% \markboth{IEEE TRANSACTIONS ON INTELLIGENT TRANSPORTATION SYSTEMS,~Vol.~X, No.~X, XXX~2024}%
% {Lin \MakeLowercase{\textit{et al.}}: eRSS-RAMP: A Rule-Adherence Motion Planner}

\maketitle

\begin{abstract}
Driving safety and responsibility determination are indispensable pieces of the puzzle for autonomous driving. They are also deeply related to the allocation of right-of-way and the determination of accident liability. Therefore, Intel/Mobileye designed the responsibility-sensitive safety (RSS) framework to further enhance the safety regulation of autonomous driving, which mathematically defines rules for autonomous vehicles (AVs) behaviors in various traffic scenarios. However, the RSS framework's rules are relatively rudimentary in certain scenarios characterized by interaction uncertainty, especially those requiring collaborative driving during emergency collision avoidance. Besides, the integration of the RSS framework with motion planning is rarely discussed in current studies. Therefore, we proposed a rule-adherence motion planner (RAMP) based on the extended RSS (eRSS) regulation for non-connected and connected AVs in merging and emergency-avoiding scenarios. The simulation results indicate that the proposed method can achieve faster and safer lane merging performance (53.0\% shorter merging length and a 73.5\% decrease in merging time), and allows for more stable steering maneuvers in emergency collision avoidance, resulting in smoother paths for ego vehicle and surrounding vehicles.
\end{abstract}


\section{Introduction}
\IEEEPARstart{T}{raffic} crashes have been increasing at an incredible rate in the past few years, reaching approximately 1.3 million injuries each year \cite{inju2021-rp}. Besides, nearly 94 \% to 96 \% of traffic accidents are attributed to human drivers according to the National Highway Transportation Safety Administration (NHTSA) \cite{Singh2015-lo}. Therefore, autonomous vehicles (AVs) have been recently developed to reduce the accident rate and the congestion of traffic, which can perceive the environment and drive safely with little or without human manipulation. Although scientific research points out that AVs are theoretically safer than human drivers \cite{Shladover2016-ey}, public concern still arises about the safety of AVs due to several fatal tragedies over the on-road tests. The Department of Motor Vehicles (DMV) from California has received over 500 AV collision reports so far \cite{AVreport2021}. Subsequently, connected autonomous vehicles (connected AVs) have been proposed to compose the cooperative intelligent transportation systems (C-ITS) by using vehicle-to-vehicle (V2V) communication, which can improve road safety to a great extent compared to the single-intelligent AVs \cite{Hafner2022-iv}.

Despite safety assurance being the most crucial task in the research of automatic driving, the ethical issues of autonomous driving also postpone the commercialization of connected AVs, e.g., accident liability. In 2017, Mobileye introduced the responsibility-sensitive safety (RSS) framework to regulate the driving behaviors of AVs by defining a set of rigorous mathematical models, interpreting the Duty of Care law \cite{Shalev-Shwartz2017-bh}. The RSS is constructed by following the five inductive principles: \romannumeral1. Do not hit someone from behind; \romannumeral2. Do not cut in recklessly; \romannumeral3. Right-of-way is given, not taken; \romannumeral4. Be careful of areas with limited visibility; \romannumeral5. If you can avoid an accident without causing another one, you must do it. Furthermore, NHTSA published a white paper to implement the RSS framework into pre-crash scenarios for light vehicles, prompting the safety standardization of AVs \cite{Shashua2018-zn}. Multiple research institutes have joined and investigated the worldwide standardization of RSS, including the Institute of Electrical and Electronics Engineers (IEEE): Standard 2846 \cite{9761121}, China ITS Industry Alliance, etc. However, the implementation of the RSS remains an unexplored territory in terms of logical foundations and formalization for interaction uncertainty scenarios, such as merging behaviors and collaborative driving for non-connected and connected AVs \cite{Hasuo2022-he, Elli2021-xf}.

In this paper, we present a novel rule-adherence motion planner with the extended RSS framework (eRSS-RAMP), enhancing collaborative planning and collision avoidance for interaction uncertainty scenarios where the intentions and actions of surrounding vehicles are unpredictable in an upcoming merging section. The key contributions of this study are briefly summarized in the following points: 
\begin{figure*}[t]
    \centering
    \includegraphics[width=0.796\hsize]{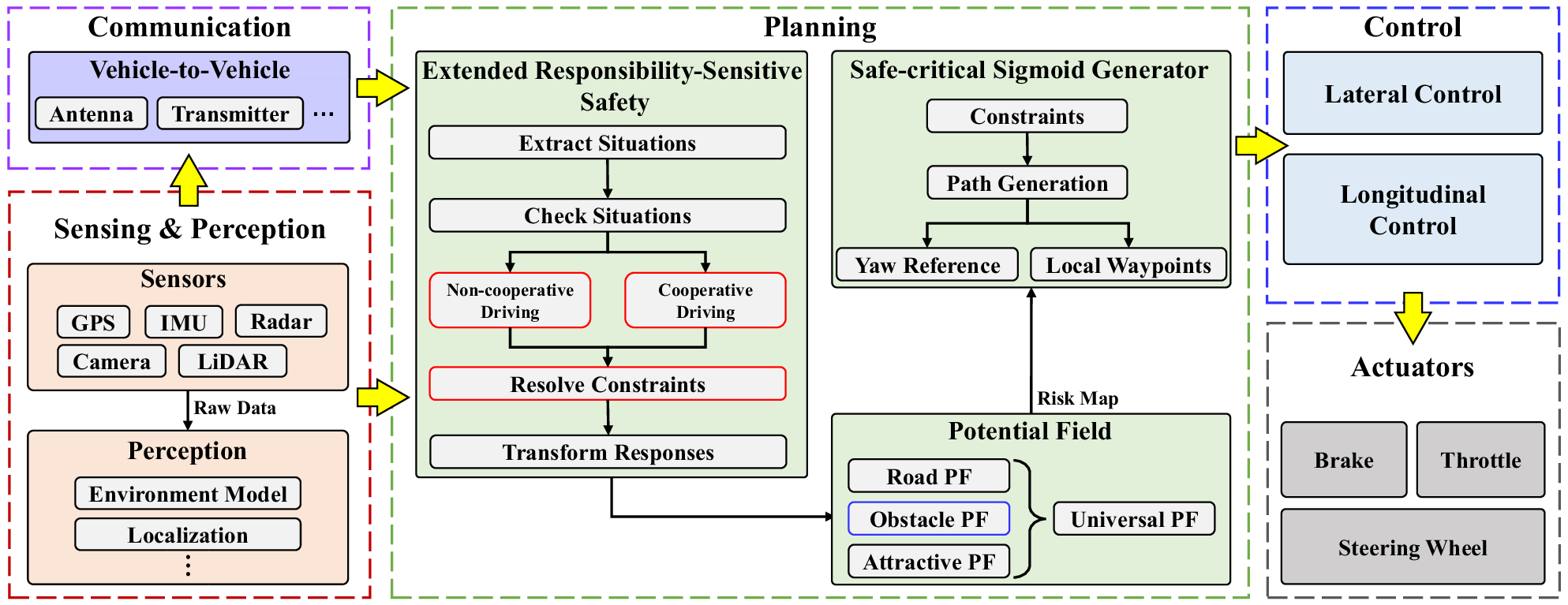}
    \caption{Overall system framework of the proposed eRSS-RAMP method: Communication components like the antenna and transmitter enable vehicle-to-vehicle interaction. Sensing \& Perception gather data from various sensors and process it for environment modeling and localization. Planning includes the extended responsibility-sensitive safety and safe-critical sigmoid generator to manage reference signals (paths, velocity, etc.), while the control module oversees lateral and longitudinal control, directing the actuators (brake, throttle, steering wheel).}
    \label{system_scheme}
\end{figure*}
\begin{itemize}
    \setlength{\itemsep}{0pt}
    \setlength{\parsep}{0pt}
    \setlength{\parskip}{0pt}
    \item We expand the RSS framework regarding several typical driving scenarios (merge and emergency avoidance), including the proper response and desired cooperative motion for non-connected and connected AVs.
    \item We design an alternative path planner using the sigmoid curve to address the local minima of the PFs. Additionally, we apply the minimum safe distances of the RSS as constraints to regulate the parameters of the sigmoid.
    \item We combine the RSS framework with the potential fields (PFs) by considering the V2V communication delay and more obstacle motion status for risk modeling.  
\end{itemize}

The overall system framework is depicted in Fig. \ref{system_scheme}, where we mainly focus on the planning module: extended RSS, PFs, and safe-critical sigmoid generator with speed planning. In addition, the RSS sub-module will enforce restrictions on the actuators' commands if the vehicle violates the safety regulations. The simulation study was evaluated using MATLAB/Simulink in merging and emergency-avoiding scenarios. The results have shown that the proposed eRSS-RAMP outperforms other methods in terms of merging efficiency (time, path length, etc.) and generating collision-free trajectories in urgent situations.

The remainder of this article is organized as follows: a detailed review of related work is presented in Section II, and Section III introduces the potential field for collision avoidance. Section IV develops the rule-adherence motion planner with extended RSS. Section V illustrates simulation results and analysis. Finally, conclusions and future discussions are provided in Section VI.

\section{Related Work}\label{related_work}

This section reviews the related work of the recent developments of the RSS framework. Toward the deployment of RSS standardization, Gassmann et al. \cite{Gassmann2019-ou, Gassmann2020-ms} developed an open-source executable library based on C++ and combined it with several autonomous driving platforms, including Carla Simulator and the Baidu Apollo project. However, it only concentrated on the software investment of longitudinal and lateral RSS minimum safe distances. Naumann et al. \cite{Naumann2019-ro} proposed to embed the RSS regulations in provable lane changes for the non-reckless driving formulation, but the detailed implementation of RSS was not specified, especially the detailed implementation in the motion planning layer. Later on, Naumann et al. \cite{Naumann2021-qw} and Konigshof et al. \cite{Konigshof2022-ot} presented a detailed approach to select proper parameters of longitudinal and lateral RSS minimum safe distances, based on the HighD dataset \cite{Krajewski2018-on}. Nevertheless, the application of the RSS framework in motion planning for AVs was not further discussed. Xu et al. \cite{Xu2021-er} presented a novel calibration and evaluation of the RSS for cut-in scenarios involving minimal time-to-collision (TTC). However, the RSS was only embedded in the adaptive cruise control (ACC) model without further improvements. Similarly, Chai et al. \cite{Chai2020-ou} introduced an improved RSS with triggering conditions when applied to ACC models, but it only evaluated the longitudinal motion of the vehicle and was not combined with any planning algorithms. Then, a reachable set analysis was introduced for the verification of RSS-based AVs by Orzechowski et al. \cite{Orzechowski2019-ku}. However, the simulation did not consider emergency scenarios; only low-speed scenarios were studied. On the other hand, Pasch et al. \cite{Pasch2021-gc} used the RSS for vulnerable road users (pedestrians mainly) in structured environments; nevertheless, the RSS was used for only longitudinal evaluation, and the interaction uncertainties between AVs were not involved. Soon afterwards, an extended version of RSS (RSS$^+$) was proposed by Oboril et al. \cite{Oboril2021-pa} to solve the potentially dangerous situations; but it did not specify the RSS combined with motion planning between connected AVs, and the RSS$^+$ could result in over-conservative driving behaviors. To reduce the over-conservative calculations of the RSS, Hassanin et al. \cite{Hassanin2022-wn} proposed a modified version (RSS$\_$X) that uses the positive previous time-step acceleration and standstill gap to replace the maximum acceleration. However, this modification only considered longitudinal controls for car-following scenarios. Similarly, to solve the dilemma of RSS in mixed traffic during longitudinal maneuvers, Qi et al. \cite{Qi2024-dz} proposed a human error-tolerant (HET) driving strategy, where AVs maintain an extra gap and prepare for moderate deceleration to compensate for human driver errors. For occlusion problems, Gassmann \cite{Gassmann2023-kb} and Lin \cite{Lin2023-tj} presented the implementation of the RSS model to address the limited visibility in autonomous driving and evaluate the balance between safety and practicality. Therefore, they did not discuss the interaction uncertainties between AVs. Recently, He et al. \cite{He2024-ao} used the RSS model to design a safety mask to ensure collision safety during the training and testing processes of reinforcement learning (RL). However, the study focused on the decision-making layer with less emphasis on motion planning. Hu et al. \cite{Hu2024-xy} also incorporated the RSS and lane-changing models with RL to enhance the safety and efficiency of autonomous driving decision-making. To accurately evaluate risk and anticipate hazards, Zhang et al. \cite{Zhang2024-uk} proposed a Psycho-Physical Field (PPF)-based Real-time Driving Risk Assessment (RDRA) framework that combines the RSS framework with the risk field using both physical and psychological factors. However, the case study only evaluated longitudinal motions (braking maneuvers).

Consequently, all the above insights motivated us to explore the broader implementation of the RSS framework, particularly for interaction uncertainty scenarios, including not only longitudinal motions but also lateral driving regulations. 

\section{Potential Field for Collision Avoidance}\label{rss_pf}

In this section, we introduce the RSS-enhanced PFs and the safe-critical sigmoid generator with speed planning for collision avoidance.
%
% \begin{figure}[t]
% % \centering
%     \subfigure[3D view of road potential field]{
%     \includegraphics[width=\hsize]{fig_tvt/road_pf_3D.eps}
%     }
%     \subfigure[2D view of road potential field along the Y-Z axis]{
%     \includegraphics[width=\hsize]{fig_tvt/road_pf_sideview.eps}
%     }
% \label{road_pf}
% \caption{Road potential field for a two-lane highway road}
% \end{figure}
% \begin{figure}[t]
%     \centering
%     \includegraphics[width=\hsize]{fig_tvt/road_pf_3D.eps}
%     \caption{Road potential field for a two-lane highway road}
%     \label{road_pf}
% \end{figure}

\subsection{Potential field}

The PFs in autonomous driving refer to a technique used for navigation and obstacle avoidance, where vehicles are guided by a virtual field that exerts attractive forces towards the destination and repulsive forces away from obstacles. This method helps in planning safe and efficient paths for AVs. It usually applies mathematical functions to build the fields according to the road components, including road structure, obstacle vehicles, pedestrians, etc. The potential functions used for road structure and pedestrians have been well-defined and explained in Ji et al. \cite{Ji2017-fk} and \cite{Raksincharoensak2016-og}. In this paper, we primarily improve the PF for obstacle vehicles by introducing the longitudinal and lateral safe distance definitions of the RSS for more precise modeling.

The obstacle potential functions (OPFs) of obstacle vehicles are designed to maintain a safe distance from the ego vehicle and to properly guide a lane-changing maneuver for the ego vehicle when necessary. Thus, the oval shape is frequently used for multiple scenarios, including lane change, overtaking, etc. The following exponential function is formulated for modeling the expected behavior \cite{Wang2019-lz}.
\begin{equation}
    \begin{split}
        P_{OV}=\sum_{j=1}^{m}
        &A_{OV} \exp[-\frac{C_1^j}{2}(\frac{(X-X_o^j)^2}{\sigma_x^j}+\\
        &\frac{(Y-Y_o^j)^2}{\sigma_y^j}-C_2^j)],
    \end{split}
    \label{obs_pf}
\end{equation}
where
\begin{align*}
    &C_1^j=1-{\psi_o^j}^2,\quad
    C_2^j=\frac{2\psi_o^j(X-X_o^j)(Y-Y_o^j)}{\sigma_x^j\sigma_y^j},\\
    &\sigma_x^j=D_{rss}^{long,j}\sqrt{-\frac{1}{\ln{\epsilon}}},\quad
    \sigma_y^j=\sqrt{-\frac{{D_{rss}^{lat,j}}^2}{2\ln{\frac{\epsilon}{A_{OV}}}}}.
\end{align*}
where $A_{OV}$ denotes the maximum amplitude of the OPFs and $X$ is the longitudinal position of the ego vehicle, $X_{o}^j,Y_{o}^j$ are the longitudinal and lateral positions of $j^{th}=1,2,\cdots,m$ obstacle vehicle. $\sigma_{x}^j, \sigma_{y}^j$ are the longitudinal and lateral coefficients of the OPFs. $\psi_o^j$ is the yaw angle of the $j^{th}$ obstacle vehicle and $\epsilon$ represents a minimum positive factor. Towards the standardization of driving safety, we introduce the minimum safe distances of the RSS criteria with concrete improvements, including the longitudinal ($D_{rss}^{long,j}$) and lateral ($D_{rss}^{lat,j}$) directions. Since \cite{Shalev-Shwartz2017-bh} has clearly described the definitions, we make a supplementary about the minimum safe distances for connected AVs by considering the communication delay and the sizes of different vehicles.
\begin{lemma}[\textbf{Safe Longitudinal Distance}] Let $c_e$ be the ego vehicle that is behind the front $j^{th}$ obstacle vehicle $c_f^j$ on the X-axis. Let $v_e$, $v_f^j$ be the longitudinal velocities of the vehicles. Then, the minimum safe distance for the X-axis between $c_e$ and $c_f^j$ is:
\begin{figure}[t]
    \centering
    \includegraphics[width=0.65\hsize]{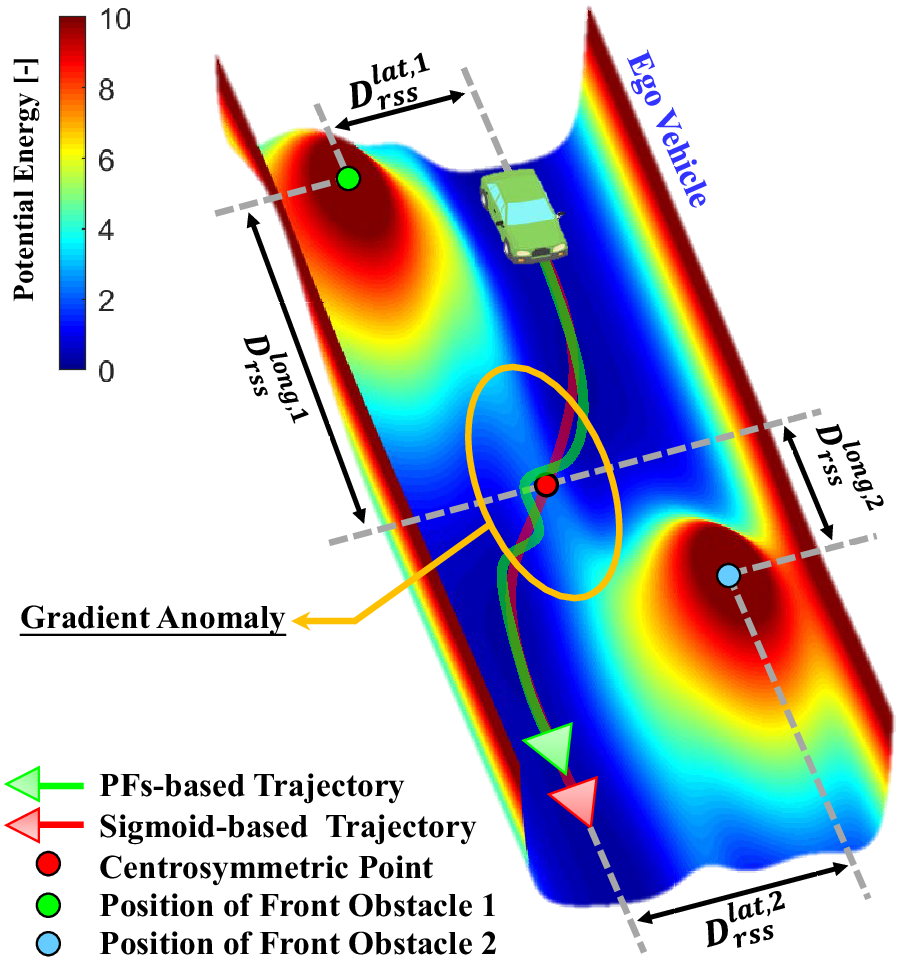}
    \caption{Sigmoid-based trajectory with the PFs (solid red line) and only PFs-based trajectory (solid green line).}
    \label{pf_rss_sig}
\end{figure}
\begin{equation}
    \begin{split} 
        D_{rss}^{long,j}=
        &[ v_e(\rho_{rt}+\rho_{cd})+\frac{1}{2}a_{accel}^{max}(\rho_{rt}+\rho_{cd})^2+\frac{l_e+l_f^j}{2}+\\
        &\frac{(v_e+ a_{accel}^{max}(\rho_{rt}+\rho_{cd}))^2}{2a_{brake}^{min}}-\frac{{v_f^j}^2}{2a_{brake}^{max,j}} ]_+,
        \label{long_rss}
    \end{split}
\end{equation}
\label{lema1}
where $[\textit{h}]_+:\max\{\textit{h},0\}$
\end{lemma}
\noindent with $\rho_{rt}$ is the reaction time of the ego vehicle in sensing the emergency and $\rho_{cd}$ is the communication delay between connected AVs; $a_{accel}^{max}$ and $a_{brake}^{min}$ denote the maximum acceleration and the minimum braking of the ego vehicle; $a_{brake}^{max,j}$ is the maximum braking of the $j^{th}$ obstacle vehicle; $l_e$ denotes the length of the ego vehicle and $l_f^j$ is the length of the $j^{th}$ obstacle vehicle. Regarding the reaction time for AVs, \cite{Dixit2016-lt} investigated the recent traffic data released from the California trials, which is approximately 0.83 seconds on average. In this study, we assume there is no packet loss during the transmission. The communication delay between connected AVs is assumed to be around 0.5 ms which is estimated from the real testing data \cite{Cao2020-ex}. On the other hand, we have the following proposition for calculating the minimum safe distance for the Y-axis.
\begin{lemma}[\textbf{Safe Lateral Distance}] Without loss of generality, assume that the ego vehicle $c_e$ is to the left of the $j^{th}$ obstacle vehicle $c_o^j$. Define $v_{e,\rho}^{lat}=v_e^{lat}+a_{accel}^{lat,max}(\rho_{rt}+\rho_{cd})$ and $v_{o,\rho}^{lat,j}=v_o^{lat,j}-a_{accel}^{lat,max,j}(\rho_{rt}+\rho_{cd})$. Then, the minimum safe distance for the Y-axis is:
\begin{equation}
    \begin{split}
        D_{rss}^{lat,j}=
        &\mu+\Bigg[\frac{v_e^{lat}+v_{e,\rho}^{lat}}{2}(\rho_{rt}+\rho_{cd})+\frac{{v_{e,\rho}^{lat}}^2}{2a_{brake}^{lat,min}}+\frac{l_w+l_{w}^j}{2}\\
        &-\left(\frac{v_{o}^{lat,j}+v_{o,\rho}^{lat,j}}{2}(\rho_{rt}+\rho_{cd})+\frac{{v_{o,\rho}^{lat,j}}^2}{2a_{brake}^{lat,min,j}}\right)\Bigg]_+,
        \label{lateral_rss}
    \end{split}
\end{equation}
\end{lemma}
\noindent where $lat$ denotes the vehicle motions in Y-axis and $\mu$ determines the fluctuation margin; $l_w$ and $l_w^j$ represent the widths of the ego vehicle and the $j^{th}$ obstacle vehicle, respectively. Eq. \ref{long_rss} and \ref{lateral_rss} indicate that the minimum safe distances of the RSS consider the worst-case collision that can lead to over-conservative driving behavior. Therefore, Orzechowski et al \cite{Naumann2021-qw} proposed reasonable parameter sets for the RSS that enable thorough safety verification without impeding traffic flow. The demonstration of the PFs and RSS combination is depicted in Fig. \ref{pf_rss_sig}. 

\subsection{Safe-critical Sigmoid Generator}
\begin{figure}[t]
    \centering
    \includegraphics[width=0.85\hsize]{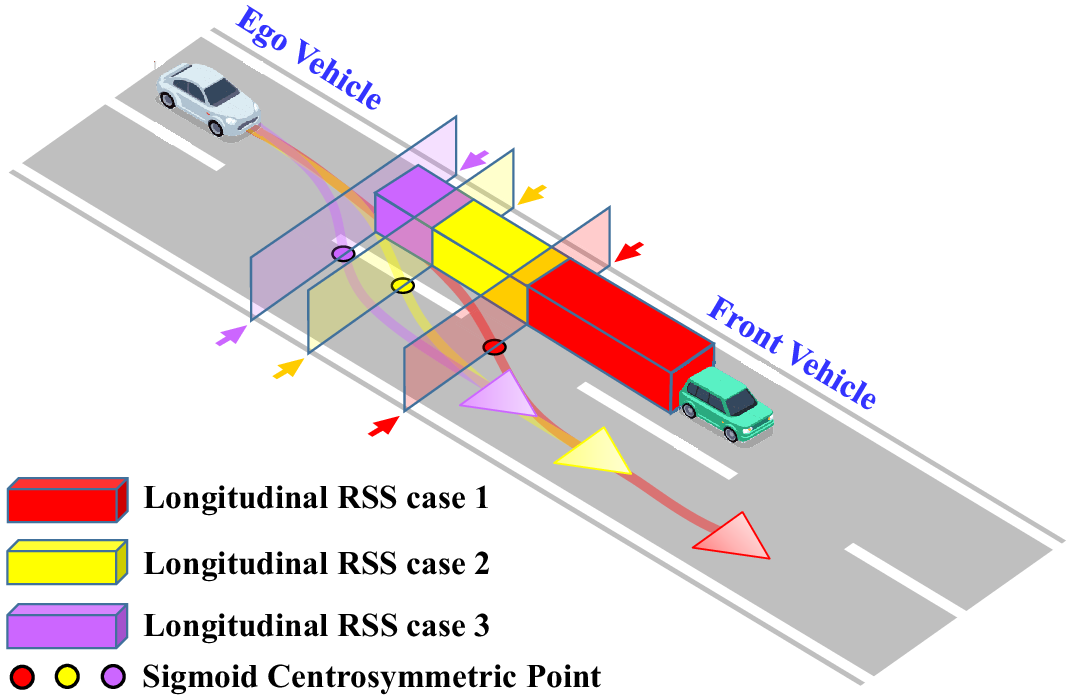}
    \caption{Slope at centrosymmetric point of the sigmoid curve.}
    \label{slope_sig}
\end{figure}
% \begin{figure}[t]
%     \centering
%     \includegraphics[width=\hsize]{fig_tvt/rss_pf_des.eps}
%     \caption{RSS-based potential field for modeling collision risk with the front obstacle vehicle, including the minimum safe distances along the X-Y coordinates.}
%     \label{pf_rss}
% \end{figure}

Yoren et al \cite{Koren1991-ar} pointed out one inherent limitation that the path will exhibit local oscillations when the obstacles' PFs are closely spaced, as shown in Fig. \ref{pf_rss_sig}. Lu et al. \cite{Lu2020-kk} proposed a hybrid path planning method that uses the sigmoid curve to generate a set of path candidates and then designed a nonlinear objective function to select the best one instead of the PF. Differing from the nonlinear optimization approach, we generate the sigmoid curve under the criterion of RSS-based PFs for safe-critical consideration and time efficiency. The standard formulation of a sigmoid function $f_{sig}$ is denoted as follows.
\begin{equation}
    f_{sig}(X)=
    \frac{W}{1+e^{-\kappa(X-P_c)}}+b,
\end{equation}
where $W$ is the lateral width of the sigmoid curve, $\kappa$ denotes the slope at the centrosymmetric point (CP) of the sigmoid curve; $P_c$ determines the longitudinal position of the CP, and $b$ denotes the bias from the X-axis. As shown in Fig. \ref{pf_rss_sig}, the selections of $W$, $P_c$, and $b$ are constrained by the following restrictions that are dependent on the road structure and the minimum safe distances of the surrounding objects.
\begin{align}
    &Y_l+\frac{l_w}{2}< b\leq \min (Y_{lane}^i, Y_o^{j+1}-D_{rss}^{lat,j+1}),\tag{10a}\\
    \max &(Y_{lane}^i,Y_o^{j}-D_{rss}^{lat,j})< W+b\leq Y_u-\frac{l_w}{2},\tag{10b}\\
    &X_o^j+D_{rss}^{long,*}\leq P_c\leq X_o^{j+1}-D_{rss}^{long,j+1}.\tag{10c}
\end{align}
It should be noted that the inequality (10c) could be untenable if $X_o^j+D_{rss}^{long,j}\geq X_o^{j+1}-D_{rss}^{long,j+1}$. In that case, the ego vehicle should revoke the lane-changing maneuver and respond properly according to the situation, which will be discussed in the next section. The final argument pertains to the slope of the sigmoid curve, which is associated with the time required for a lane change and the path curvature for riding comfort. Though Cesari et al. \cite{Cesari2017-jx} point out that it is difficult to find an explicit mathematical function for $\kappa$, we can still narrow the range for selecting it by introducing the RSS constraints, as indicated below.
\begin{equation}
    \kappa_{con}^{min}\leq \kappa \leq \max (\frac{D_{rss}^{long,j}}{X_{o}^{j}-X}, k_{con}^{max}),
\end{equation}
where $\kappa_{con}^{min}$ and $\kappa_{con}^{max}$ denote the minimum and maximum slope of the sigmoid curve, respectively. As depicted in Fig. \ref{slope_sig}, the slope of the sigmoid curve is primarily determined by the ratio of the longitudinal safe distance to the relative distance between the ego vehicle and the front $j^{th}$ obstacle vehicle. It is worth noting that the larger the slope, the shorter the lane change time, but the ride comfort will also be reduced. Additionally, we define $P_c=X_o^j-D_{rss}^{long,j}$ when the obstacle vehicles only appear ahead of the ego vehicle within the range of perception.

\section{Rule-Adherence Motion Planner with Extended Responsibility-Sensitive Safety}

In this section, we present the proposed rule-adherence motion planner approach with the extended RSS, which includes regulated proper responses (desired cooperative motions: steering, braking, and accelerating) and backup plans (in case other AVs are either non-cooperative or do not respond).
\begin{figure*}[t]
\centering
    \subfigure[Scenario 1: Non-cooperative for ahead merging]{
    \begin{minipage}[t]{0.368\linewidth}
    \centering
    \includegraphics[width=\hsize]{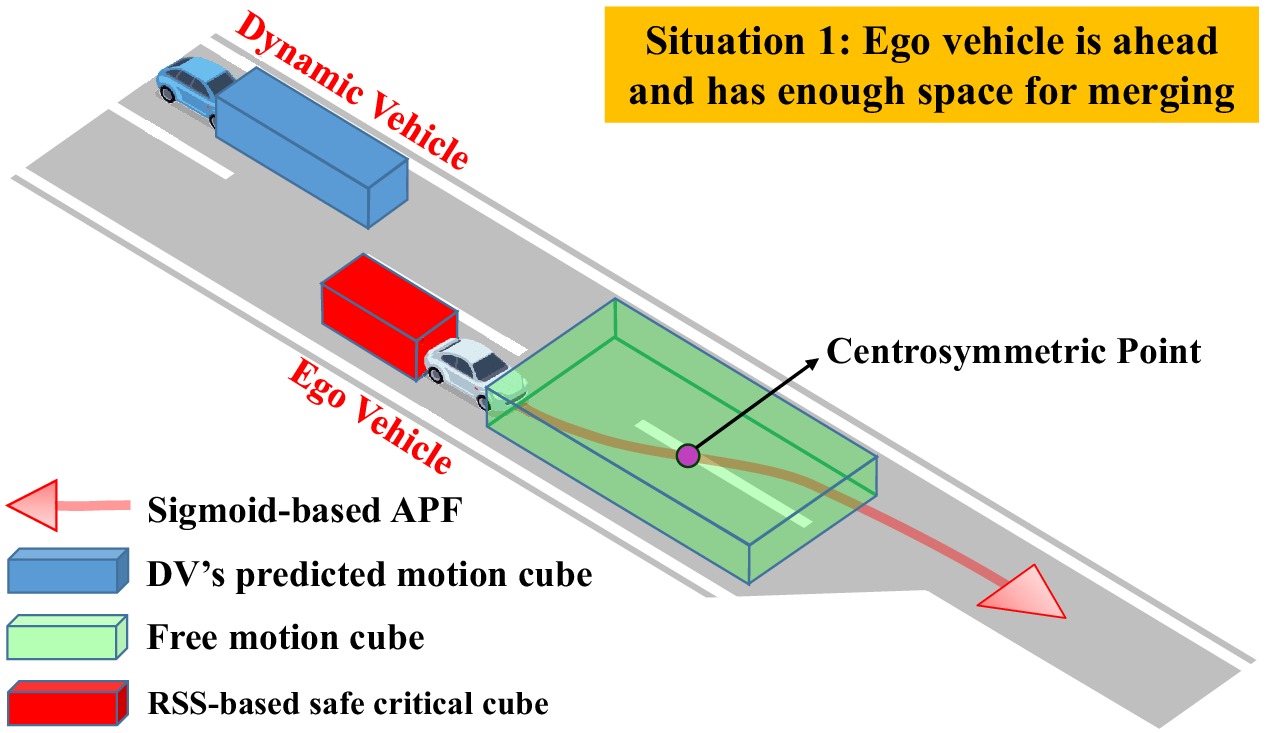}
    \end{minipage}
    \label{mer_s1}
    }%
    \subfigure[Scenario 2: Non-cooperative for behind merging]{
    \begin{minipage}[t]{0.368\linewidth}
    \centering
    \includegraphics[width=\hsize]{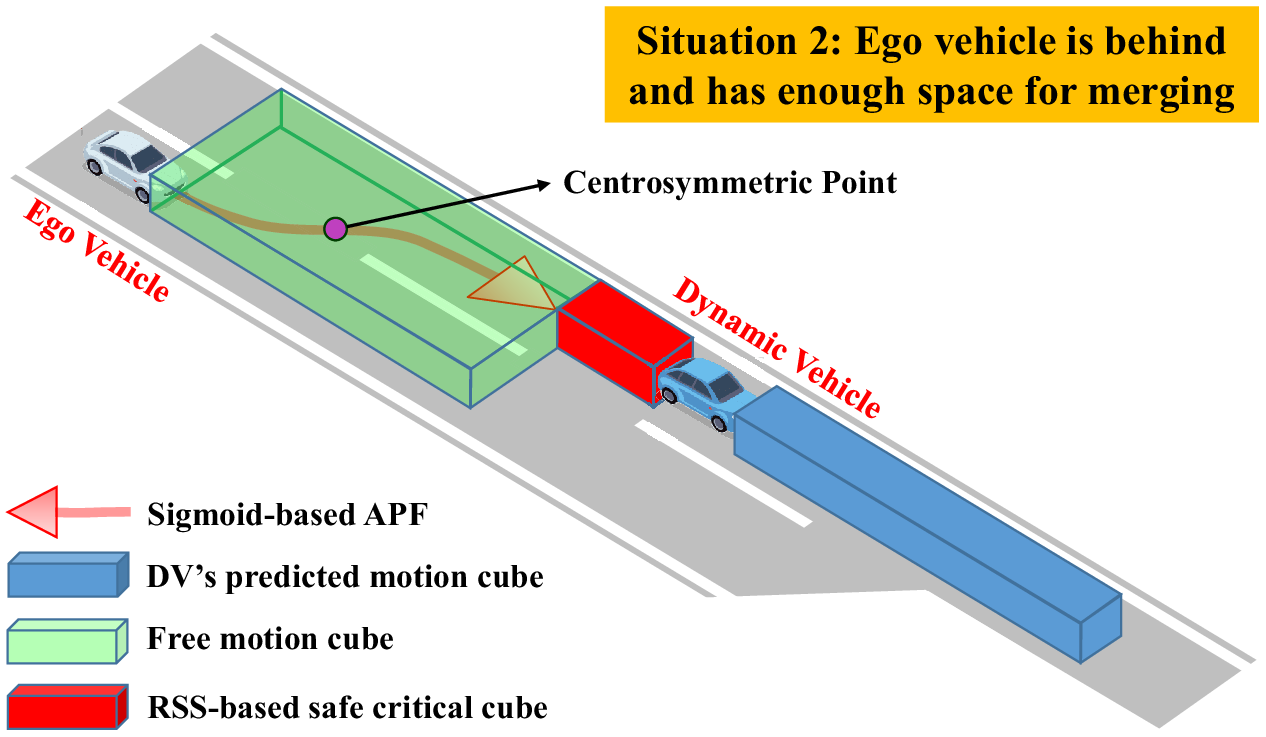}
    \end{minipage}
    \label{mer_s2}
    }
    
    \subfigure[Scenario 3: Cooperative for ahead merging]{
    \begin{minipage}[t]{0.368\linewidth}
    \centering
    \includegraphics[width=\hsize]{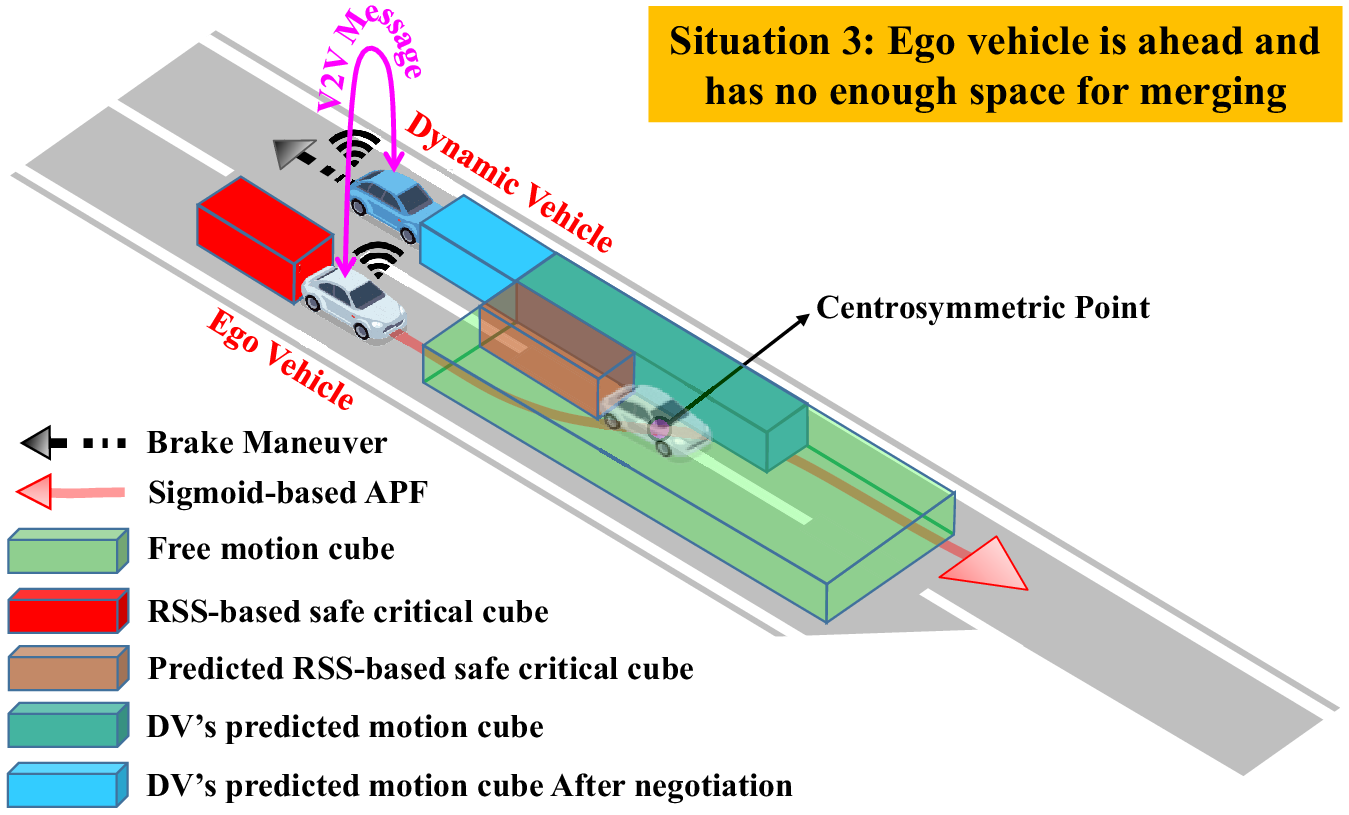}
    \end{minipage}
    \label{mer_s3}
    }%
    \subfigure[Scenario 4: Cooperative for behind merging]{
    \begin{minipage}[t]{0.368\linewidth}
    \centering
    \includegraphics[width=\hsize]{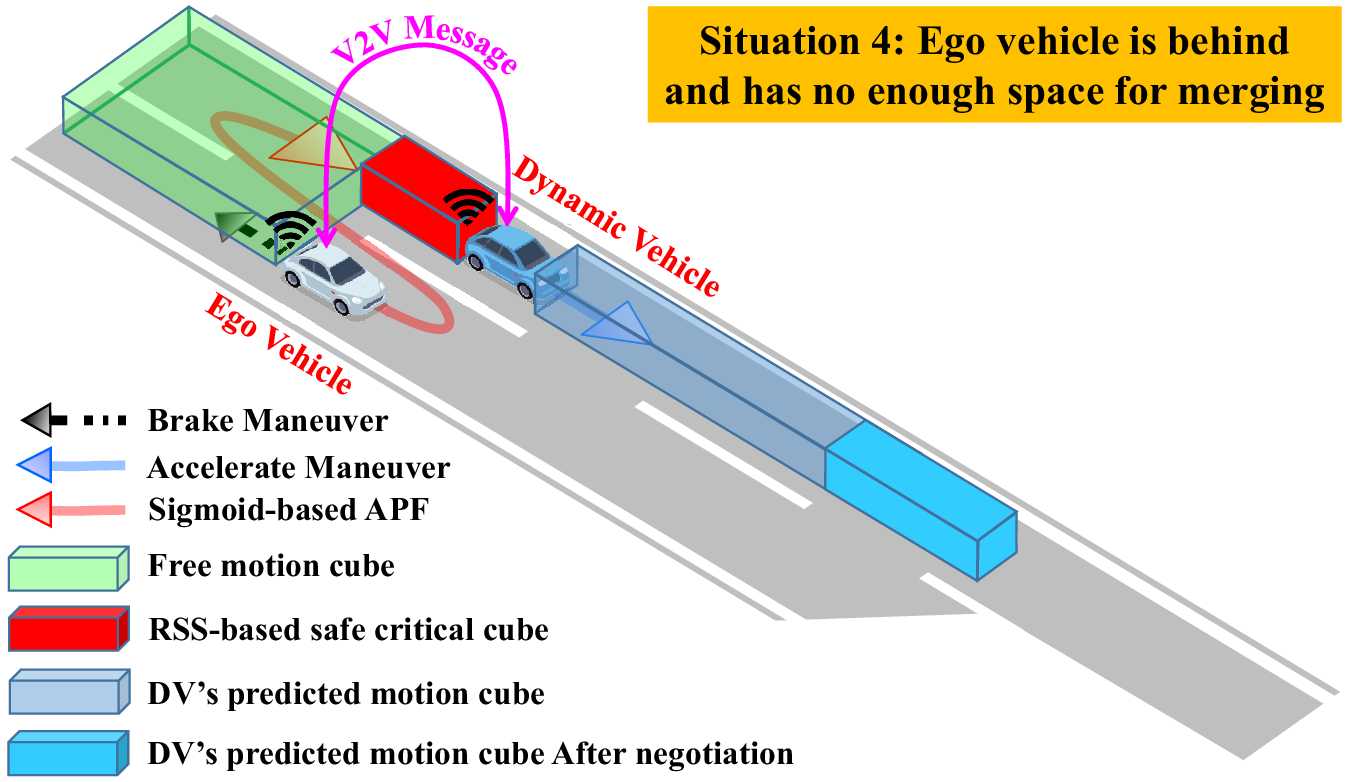}
    \end{minipage}
    \label{mer_s4}
    }
\caption{Merging Scenarios with different conditions of the traffic}
\label{mer_scen}
\end{figure*}

\subsection{Lane Merging on the Highway}

In the white paper outlining the RSS framework \cite{Shalev-Shwartz2017-bh}, the authors provide a brief overview of road conditions involving multiple route geometries, such as roundabouts and highway merges, emphasizing the need to adjust definitions of safe distances and proper responses accordingly for complex scenarios to ensure appropriate allocation of right-of-way. However, this white paper only defines longitudinal ordering for two routes in a broad sense. It does not elaborate on how to specifically adjust safe distances or detail the correct lateral and longitudinal responses. Therefore, we provide more specific definitions for highway merges, including safety specifications, proper responses, etc. Additionally, we also considered the willingness of neighboring vehicles to cooperate (cooperative or non-cooperative) and their response delays. As depicted in Fig. \ref{mer_scen}, the four merging scenarios can cover the vast majority of driving situations, taking into account both near and far relative distances for front and rear highway merges.  Here is the first revised definition to regulate the situations referred to in Figs. \ref{mer_s1} and \ref{mer_s2} for non-connected AVs.
\begin{myDef}[\textbf{Merging in Non-cooperative Driving}] Let $c_o^j$ be the $j^{th}$ obstacle vehicle on the main lane. Let $c_e$ be the ego vehicle on the side lane that is merged into the main lane. Let $\rho_{lc}$ be the lane-change threshold time, and let $T_{LC}^{dec}$ be the lane-change decision time. The merging maneuver is safe for the two non-cooperative vehicles that drive in the same direction by complying with the following constraints:

    \textbf{1.} If at the interval $[T_{LC}^{dec},\;T_{LC}^{dec}+\frac{\rho_{lc}}{2}]$, and say that $c_e$ is ahead $c_o^j$; then:
    \begin{equation}
        X_{o}^j+\frac{v_o^j\rho_{lc}}{2}+\frac{a_{accel}^{max,j}\rho_{lc}^2}{8}\leq X+v_e^*\frac{\rho_{lc}}{2}-D_{rss}^{long}
        \label{ineq_1}
    \end{equation}
    
    \textbf{2.} If at the interval $[T_{LC}^{dec},\;T_{LC}^{dec}+\rho_{lc}]$, and say that $c_e$ is behind $c_o^j$; then:
    \begin{equation}
        X+v_e^*\rho_{lc}\leq X_{o}^j+v_o^j\rho_{lc}-D_{rss}^{long,j}
        \label{ineq_2}
    \end{equation}
    
    \textbf{3.} If there is continuously insufficient space for $c_e$ to merge into, $c_e$ must stop completely before reaching the end of the side lane, seizing for re-merge.
\label{def1}
% \vspace{-0.66cm}
\end{myDef}
\noindent where ``cubes'' represent different spatial regions that must be considered during vehicle merging operations, taking into account the vehicle's dimensions and other safety specifications. In the first constraint of Definition \ref{def1}, we consider the worst case that the rear vehicle $c_o^j$ must accelerate at most $a_{accel}^{max,j}$ during the time interval $[T_{LC}^{dec},\;T_{LC}^{dec}+\frac{\rho_{lc}}{2}]$. After $T_{LC}^{dec}+\frac{\rho_{lc}}{2}$, $c_o^j$ will brake properly according to the first rule of the RSS when $c_e$ crosses the lane divider. On the second constraint of Definition \ref{def1}, the ego vehicle should maintain the speed during the lane-changing process according to the empirical driving suggestions \cite{8step2022}. Furthermore, $v_e^*$ indicates that the speed should be adjusted (accelerate or brake) based on the speeds of the traffic in the main lane ($v_o^{j-1}$ and $v_o^{j}$) before the lane change. 
\begin{figure*}[t]
\centering
\subfigure[Scenario 1: Static and closely-spaced obstacle vehicles]{
\begin{minipage}[t]{0.368\linewidth}
\includegraphics[width=\hsize]{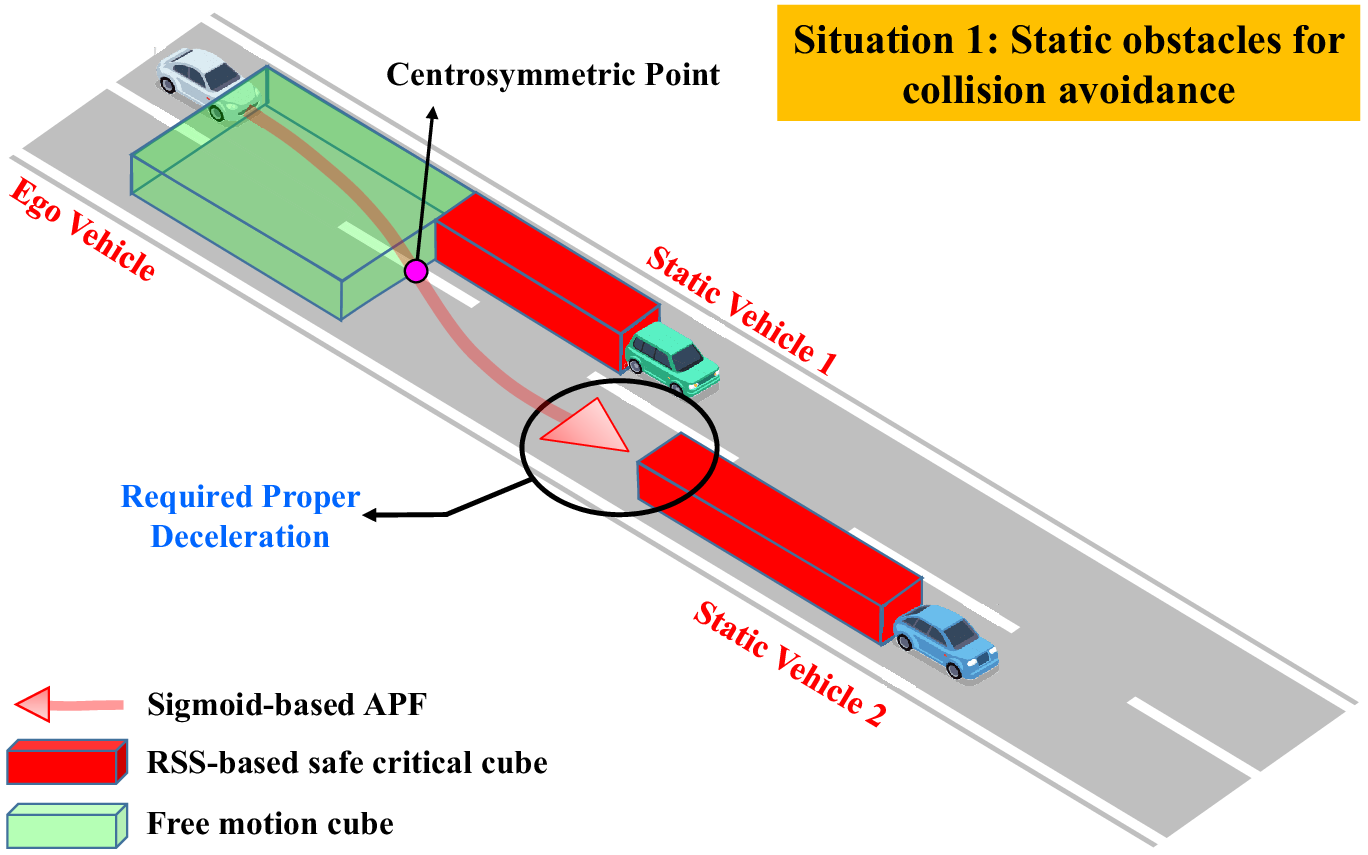}
\end{minipage}
\label{emer_s1}
}%
\subfigure[Scenario 2: Non-cooperative collision avoidance with vehicle platoon]{
\begin{minipage}[t]{0.368\linewidth}
\includegraphics[width=\hsize]{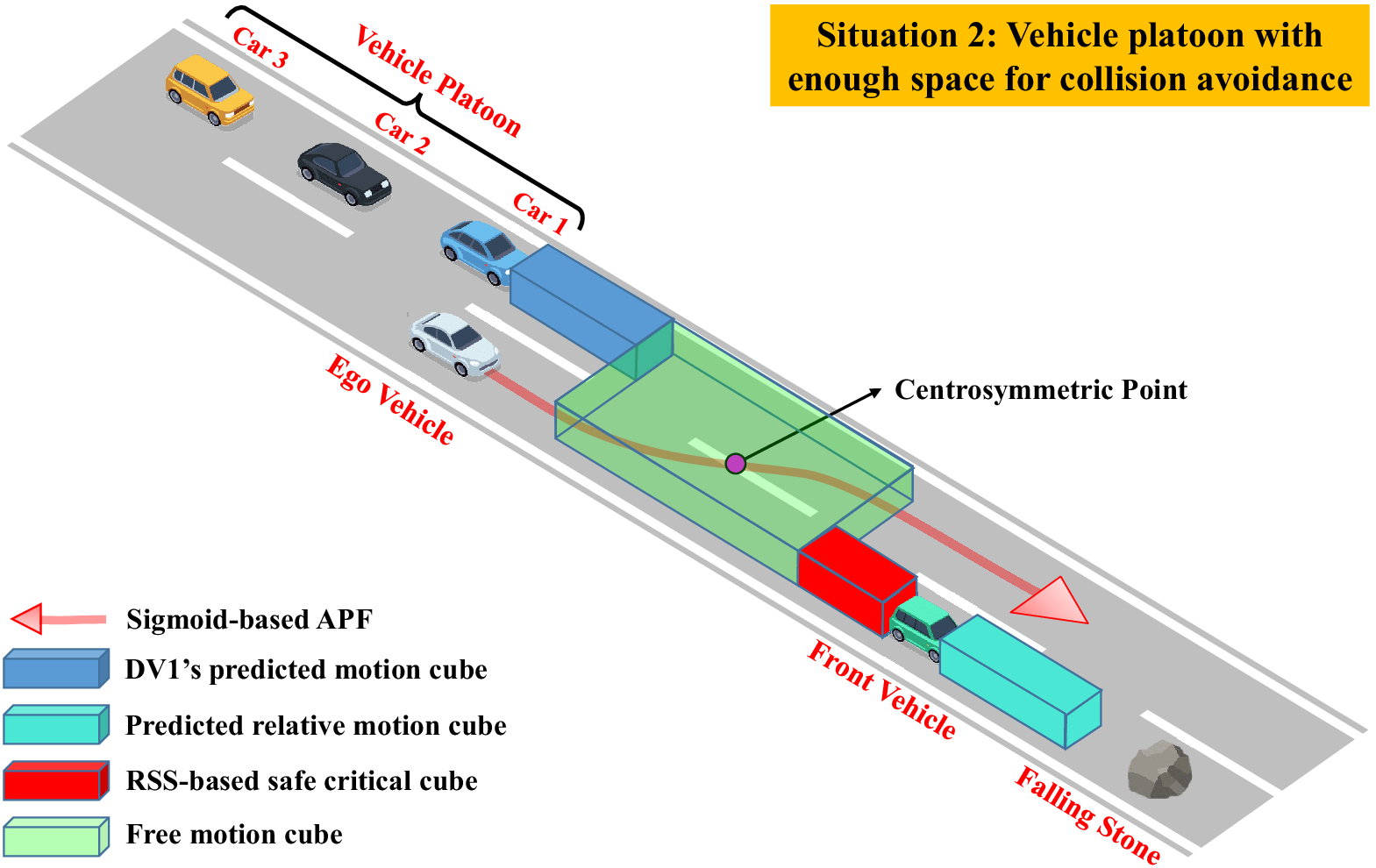}
\end{minipage}
\label{emer_s2}
}

\subfigure[Scenario 3: Cooperative collision avoidance with vehicle platoon for straight-ahead obstacle]{
\begin{minipage}[t]{0.368\linewidth}
\includegraphics[width=\hsize]{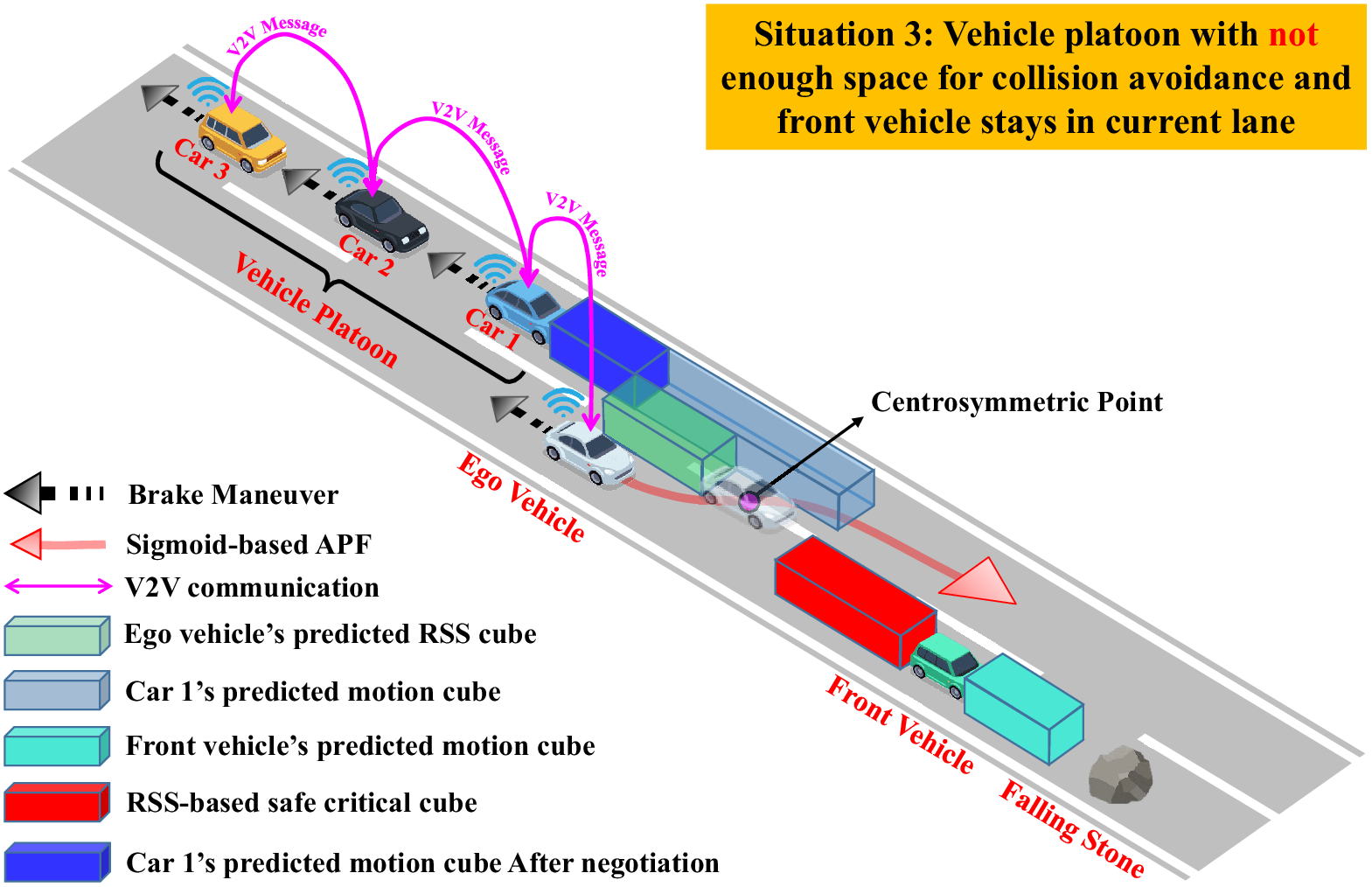}
\end{minipage}
\label{emer_s3}
}%
\subfigure[Scenario 4: Cooperative collision avoidance with vehicle platoon for lane-changing obstacle]{
\begin{minipage}[t]{0.368\linewidth}
\includegraphics[width=\hsize]{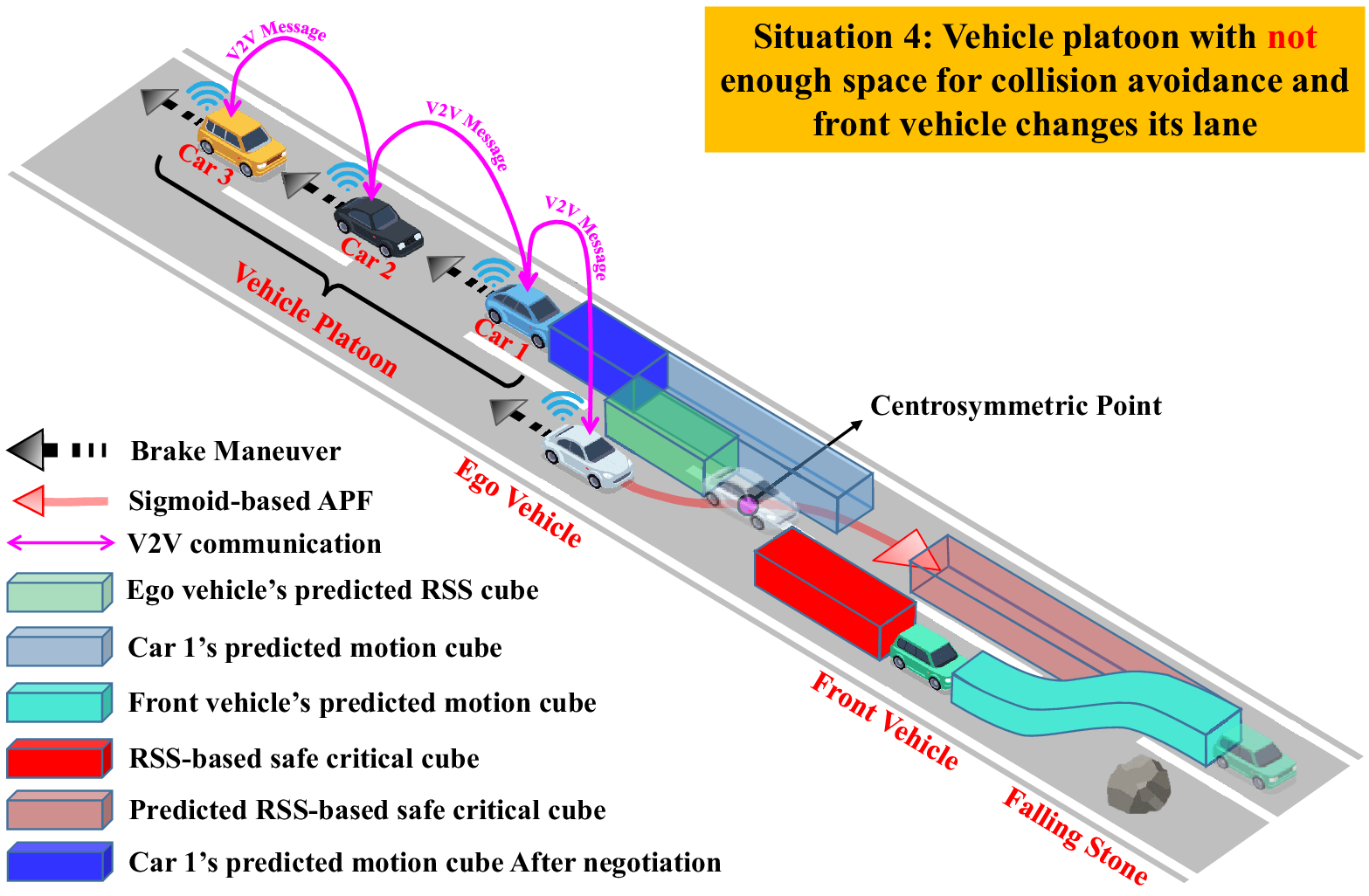}
\end{minipage}
\label{emer_s4}
}
\caption{Collision avoidance in the static environment and emergency scenarios}
\end{figure*}

On the other hand, cooperative driving is needed if the merged space is insufficient for the requirements of Definition \ref{def1}. For human drivers, the cooperative merging maneuver is conducted by using blinkers and horns. In terms of connected AVs, a more convenient way for negotiation can be achieved via vehicle-to-vehicle (V2V) communication by sharing the desired collaborative messages, including path and speed. As depicted in Figs. \ref{mer_s3} and \ref{mer_s4}, we define the following definitions for the cooperative merging maneuver.  
\begin{myDef}[\textbf{Merging in Cooperative Driving}] Let $c_e$,\; $c_o^j$,\;$T_{LC}^{dec}$ be as in Definition \ref{def1}. Let $D_{rss}^{long,*}$ be the predicted safe longitudinal distance of $c_e$ at the CP of the sigmoid curve, and let $\rho_{ct}$ be the communication threshold time. The merging maneuver is safe for the two cooperative vehicles that drive in the same direction by complying the following responses:
    \textbf{1.} If at the interval $[T_{LC}^{dec},\;T_{LC}^{dec}+\rho_{ct}]$, and say that $c_e$ is ahead $c_o^j$; then:
    \begin{enumerate}
        \item[A.] $c_e$ must send $c_o^j$ a standard V2V message with $P_c$ and $D_{rss}^{long,*}$ included.
        \item[B.] $c_o^j$ must brake at most $a_{brake}^{min,j}$ until reaching $c_o^{j,*}$ with the given constraint:
        \begin{small}
            \begin{equation}   
                X_o^j+\frac{v_o^j+v_o^{j,*}}{2}\rho_{ct}+v_o^{j,*}\frac{\rho_{lc}}{2}\!\leq\! P_c-D_{rss}^{long,*}
                \label{ineq_3}% X+v_e\rho_{ct}+
            \end{equation}
        \end{small} 
    \end{enumerate}

    \textbf{2.} If at the interval $[T_{LC}^{dec},\;T_{LC}^{dec}+\rho_{ct}]$, and say that $c_e$ is behind $c_o^j$; then:
    \begin{enumerate}
        \item[A.] $c_e$ must send $c_o^j$ a standard V2V message, and then brake at most $a_{brake}^{min}$
        \item[B.] $c_o^j$ must accelerate at most $a_{accel}^{max,j}$ until reach $v_o^{j,*}$ with the given constraint:
        \begin{small}
            \begin{equation}   
                X+v_e\rho_{ct}-\frac{a_{brake}^{min}{\rho_{ct}}^2}{2}\leq X_o^j+\frac{v_o^{j,*}+v_o^j}{2}\rho_{ct}-D_{rss}^{long,j}
                \label{ineq_4}
            \end{equation}
        \end{small}
    \end{enumerate}
    
    \textbf{3.} If at the interval $[T_{LC}^{dec},\;T_{LC}^{dec}+\rho_{ct}]$, and say that $c_o^j$ is non-cooperative or non-responsive to $c_e$; then $c_e$ must merge by following Definition \ref{def1}.
\label{def2}
\end{myDef}
\noindent Note that $v_o^{j,*}$ should also be obtained by following the RSS restrictions from $(j+1)^{th}$ obstacle vehicle while not exceeding the road speed limit. For speed planning, the speed threshold value for the ego vehicle ($v_e^*$) can be determined by solving inequalities \ref{ineq_1} and \ref{ineq_2} in non-cooperative scenarios. In cooperative scenarios, we assume that the ego vehicle will seek cooperation at the current speed ($v_e$). Therefore, the desired cooperative speed ($v_o^{j,*}$) for the $j^{th}$ obstacle vehicle  can be obtained by conforming to inequalities \ref{ineq_3} and \ref{ineq_4}. If the obstacle vehicle $c_o^j$ is either non-cooperative or non-responsive, the ego vehicle should revert to the non-cooperative merging procedures under the criteria of Definition \ref{def1}. 

\subsection{Collision Avoidance with Static Objects on the Highway}

In PF-related approaches, scenario studies typically assume that the ego vehicle overtakes static obstacles at a constant speed or by steering while braking simultaneously \cite{Rasekhipour2017-yh, Huang2020-fw}. However, maintaining high speed when passing through closely spaced obstacles is unrealistic, especially for PFs, due to the gradient anomaly \cite{Lin2022-op, Lin2022-st}. Moreover, \cite{Eckert2011-gt} points out that steering maneuvers accompanied by braking or acceleration increase the risk of losing control of the vehicle, potentially leading to a rollover. As depicted in Fig. \ref{emer_s1}, two closely spaced obstacle vehicles are parked on the highway, and the ego vehicle is driving at high speed. In this case, proper deceleration, as indicated by the black circle, is necessary under the RSS restrictions, which are outlined in the following definition.
\begin{myDef}[\textbf{Collision Avoidance with Static Objects}] The collision avoidance with static objects is safe by complying with the following constraint:
\begin{equation}
    D_{rss}^{long,j+1}\leq \frac{X_o^{j+1}-X_o^j}{2}
\label{ineq_5}
\end{equation}
\label{def3}
\end{myDef}
\vspace{-1cm}
% \noindent Hence, we can get the desired deceleration speed for the ego vehicle by solving the inequality \ref{ineq_5}.

\subsection{Collision Avoidance with Dynamic Objects on the Highway}

In collision avoidance with dynamic objects on the highway, we primarily focus on emergency scenarios where collisions are likely to occur due to unexpected events, such as falling stones. This study explores cooperative driving among AVs according to the availability of vehicle communication, as indicated in Fig. \ref{emer_s2}. We assume that the vehicle in front may suddenly engage in emergency braking and/or steering without prior warning. Similar to the non-cooperative definition of merging scenarios, we establish the following regulations for cooperative driving in emergency collision avoidance.
\begin{myDef}[Emergency Avoidance in Non-cooperative Driving] Let $c_e$,\;$\rho_{lc}$ be as Definition \ref{def1}, and let $D_{rss}^{long,*}$ be as Definition \ref{def2}. Let $c_{fv}$ be the vehicle in the same lane ahead $c_e$. Let $c_p^j$ be the $j^{th}$ vehicle of the platoon in the adjacent lane, and let $1$,\;$n$ be the index of the first and last vehicles of $c_p^j$, respectively. Let $T_{DT}^{long}$ be the longitudinal dangerous time instant, and let $\rho_{rt}$ be as Lemma \ref{lema1}. The emergency collision avoidance is safe by complying with the following responses:

    \textbf{1.} If at the interval $[T_{DT}^{long}+\rho_{rt},\;+\infty]$, $c_{fv}$ stays in current lane for braking avoidance; then: 
    \begin{enumerate}
        \item[A.] Regard $c_e$ is ahead $c_p^1$, $c_e$ must brake at most $a_{brake}^{min}$ until reaching a safe longitudinal situation. After that, the lane change maneuver is safe with the given constraint:
        \begin{equation}
            X+v_e^*\frac{\rho_{lc}}{2}-D_{rss}^{long,*}\geq X_p^j+\frac{v_p^1}{2}+\frac{a_{accel}^{long,1}\rho_{lc}^2}{8} 
        \label{ineq_6}
        \end{equation}
        \item[B.] Regard $c_e$ is behind or equal to $c_p^1$, $c_e$ must brake at least $a_{brake}^{max}$ until reaching a safe longitudinal situation, even a complete stop if necessary. After that, the lane change maneuver is safe with the given constraint:
        \begin{equation}
            X+v_e^*\rho_{lc}\leq X_p^n+v_p^n\rho_{lc}-D_{rss}^{long,n}
        \label{ineq_7}
        \end{equation}
    \end{enumerate}

    \textbf{2.} If at the interval $[T_{DT}^{long}+\rho_{rt},\;+\infty]$, $c_{fv}$ switches to the adjacent lane for steering avoidance; then:
    \begin{enumerate}
        \item[A.] Regard $c_e$ is ahead $c_p^1$, $c_e$ must brake at least $a_{brake}^{max}$ until reaching a safe longitudinal situation. After that, the lane change maneuver is safe with the given constraints of \ref{ineq_6} and below:
        \begin{align}
            X+v_e^*\rho_{lc}\leq X_{fv}+v_{fv}\rho_{lc}-D_{rss}^{long,fv}
        \label{ineq_8}
        \end{align}
        \item[B.] Regard $c_e$ is behind $c_p^j$, $c_e$ must follow Definition \ref{def4}.1.B. 
    \end{enumerate}
\label{def4}    
\end{myDef}
\noindent Definition \ref{def4} indicates that the responses for emergency collision avoidance are specified according to the collision behaviors of the vehicle ahead. After the reaction time $\rho_{rt}$, if $c_{fv}$ remains in the current lane for braking avoidance, $c_e$ should first maintain longitudinal safety and then decide on a lateral maneuver based on the adjacent vehicles $c_p^j$ within the given constraints. Otherwise, if $c_{fv}$ steers into the adjacent lane, then $c_e$ should monitor both $c_{fv}$ and $c_p^j$ while maintaining longitudinal safety. In non-cooperative situations, it is considered safer to change lanes ahead of $c_p^1$ or behind $c_p^n$ (where $n$ denotes the tail vehicle) rather than cutting in between the vehicle platoon. Therefore, using inequalities \ref{ineq_6}, \ref{ineq_7}, and \ref{ineq_8}, we can determine the desired speed threshold $v_e^*$ for the ego vehicle in non-cooperative situations. Subsequently, we introduce the distinct specifications for cooperative emergency avoidance, as shown in Figs. \ref{emer_s3} and \ref{emer_s4}, where the ego vehicle can send a V2V message to adjacent vehicles, including the desired cooperative motions (e.g., speed). Another significant consideration for cooperative motion planning is that it is safer not to stop at the accident site for the vehicle driving on the highway because it has a high probability of causing secondary accidents \cite{Sato2019-zz}. Thus, we define cooperative emergency avoidance as follows.
\begin{myDef}[Emergency Avoidance in Cooperative Driving] Let $c_e$,\;$\rho_{rt}$,\;$D_{rss}^{long,*}$,\;$c_{fv}$,\;$c_p^j(1,n)$,\;$T_{DT}^{long}$ be as Definition \ref{def4}, and let $\rho_{ct}$ be as Definition \ref{def2}. Let $\tau$ be the time headway in $c_p^j$, and let $D_{p,safe}^{min}$ be the minimum safe distance between $c_p^j$. The cooperative emergency avoidance is safe by complying with the following responses:

    \textbf{1.} If at the interval $[T_{DT}^{long}+\rho_{rt},\;T_{DT}^{long}+\rho_{rt}+\rho_{ct}]$, $c_{fv}$ stays in current lane for braking avoidance; then:
    \begin{enumerate}
        \item[A.] Regard $c_e$ is ahead $c_p^1$, $c_e$ must brake at most $a_{brake}^{min}$ until reaching a safe longitudinal situation and send $c_p^1$ a standard V2V message with $P_c$ and $D_{rss}^{long,*}$ included. After that, $c_p^1$ must brake at $a_{brake}^{min,1}$ until reaching $v_p^{1,*}$ with the given constraint:
        \begin{equation}
            \begin{split}
                P_c-D_{rss}^{long,*}-&(X_p^1+\frac{v_p^1+v_p^{1,*}}{2}\rho_{ct})\\&\geq v_p^1\tau+D_{p,safe}^{min}+\frac{l_e+l_p^1}{2}
            \label{ineq_9}
            \end{split}
        \end{equation}
        \item[B.] Regard $c_e$ is behind $c_p^{j-1}$ but ahead $c_p^j$, and say that $j\in(1,\;n]$; then $c_e$ must brake at most $a_{brake}^{min}$ until reaching a safe longitudinal situation and send standard V2V message to $c_p^j$ and $c_p^{j-1}$. After that, $c_p^j$ must follow Definition \ref{def5}.1.A, and $c_p^{j-1}$ must accelerate at most $a_{accel}^{max,j-1}$ to create safe inter-vehicle spacing for $c_e$ to change lanes under the given constraint:
        % \begin{equation}
        %     \begin{split}
        %         P_c-D_{rss}^{long,*}-&(X_p^j+\frac{v_p^j+v_p^{j,*}}{2}\rho_{ct})\\&\geq v_p^j\tau+D_{p,safe}^{min}+\frac{l_e+l_p^j}{2},
        %     \end{split}
        % \label{ineq_10}
        % \end{equation}
        % %
        \begin{equation}
            \begin{split}
                X_p^{j-1}+\frac{v_p^{j-1}+v_p^{j}}{2}&\rho_{ct}-(P_c-D_{rss}^{long,*})\\&\geq v_e\tau+D_{p,safe}^{min}+\frac{l_e+l_p^{j-1}}{2},
            \end{split}
        \label{ineq_11}
        \end{equation}
        \item[C.] Regard $c_e$ is behind $c_p^n$, $c_e$ must brake at most $a_{brake}^{min}$ until reaching a safe longitudinal situation and send a standard V2V message to $c_p^n$. After that, $c_p^n$ must accelerate at most $a_{accel}^{max,n}$ with the given constraint:
        \begin{equation}
           \begin{split}
               X_p^n+\frac{v_p^n+v_p^{n,*}}{2}&\rho_{ct}-(P_c-D_{rss}^{long,*})\\&\geq v_e\tau+D_{p,safe}^{min}+\frac{l_e+l_p^n}{2}
           \end{split} 
        \label{ineq_12}
        \end{equation}
    \end{enumerate}

    \textbf{2.} If at the interval $[T_{DT}^{long}+\rho_{rt},\;T_{DT}^{long}+\rho_{rt}+\rho_{ct}]$, $c_{fv}$ switches to the adjacent lane for steering avoidance; then:
    \begin{enumerate}
        \item[A.] Regard $c_e$ is ahead $c_p^1$, $c_e$ must follow Definition 5.1.A with the given constraint (\ref{ineq_9}) and below:
        \begin{equation}
            X+v_e\rho_{ct}+P_c\leq X_{fv}+v_{fv}(\rho_{ct}+\frac{\rho_{lc}}{2})-D_{rss}^{long,fv}
        \label{ineq_13}
        \end{equation}
        \item[B.] Regard $c_e$ is behind or equal to $c_p^1$; then $c_e$ must follow Definition \ref{def4}.1.B.
    \end{enumerate}
\label{def5}
\end{myDef}
\begin{figure*}[t]
    \centering
    \includegraphics[width=0.77\hsize]{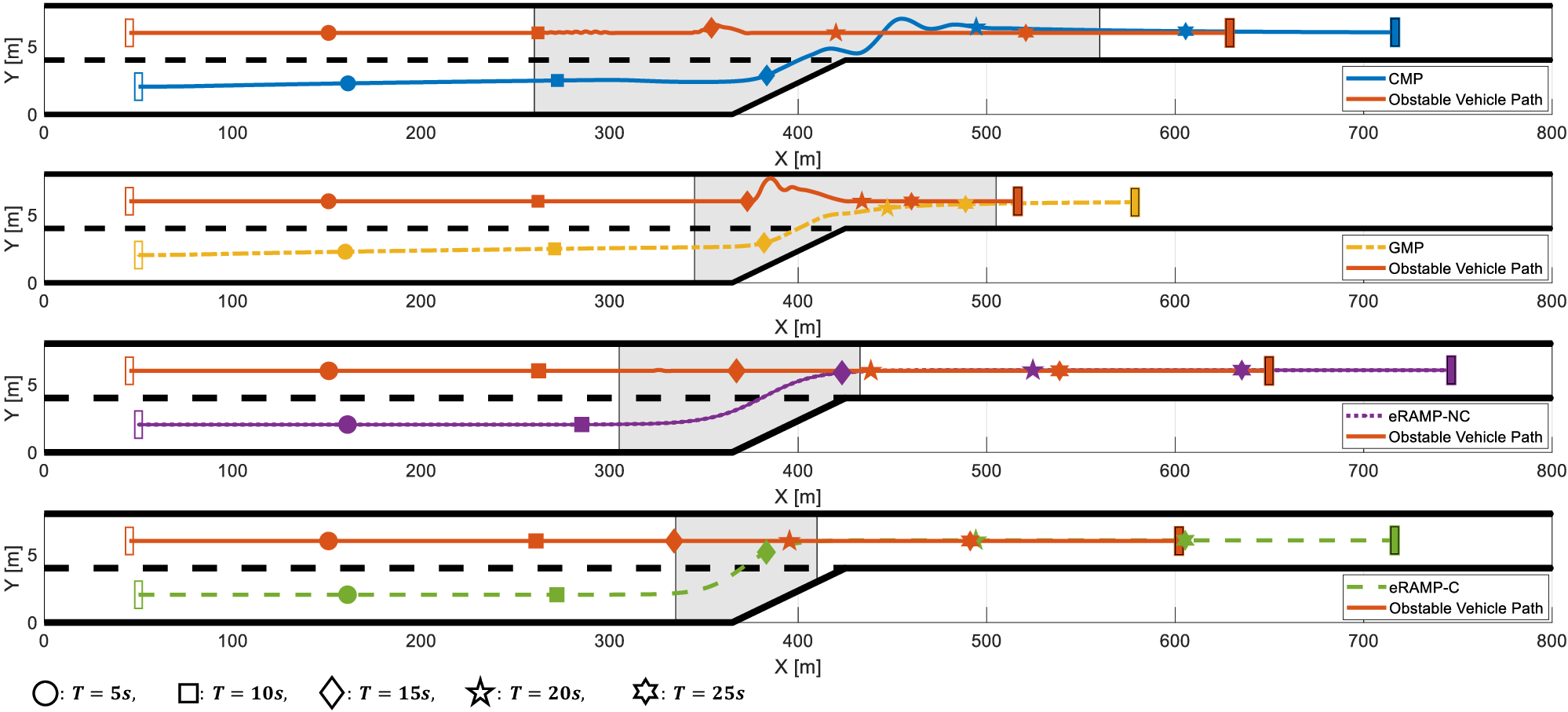}
    \caption{Merging trajectories of the ego and obstacle vehicles based on four planners: CMP, GMP, eRAMP-NC, and eRAMP-C.}
    \label{path}
\end{figure*}
\begin{figure*}[t]
\centering
    \begin{minipage}[t]{0.48\textwidth}
        \includegraphics[width=\hsize]{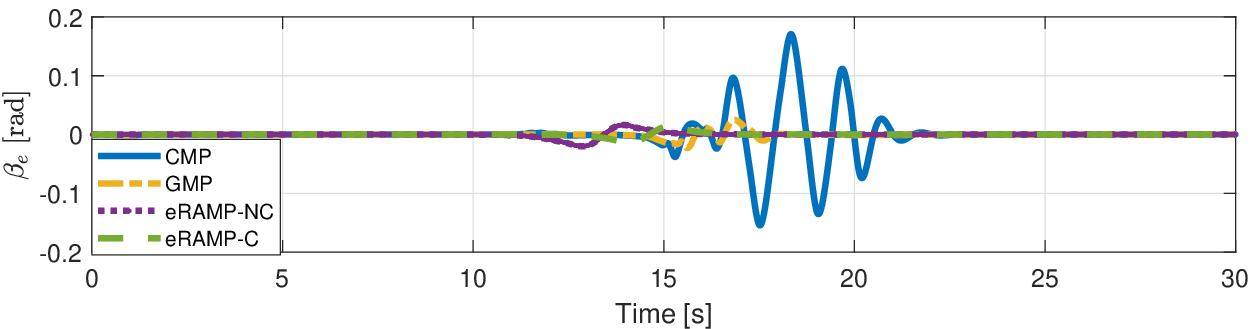}
        \caption{Sideslip angles of the ego vehicle.}
        \label{beta}
    \end{minipage}\hspace{0.15in}
    \begin{minipage}[t]{0.48\textwidth}
        \includegraphics[width=\hsize]{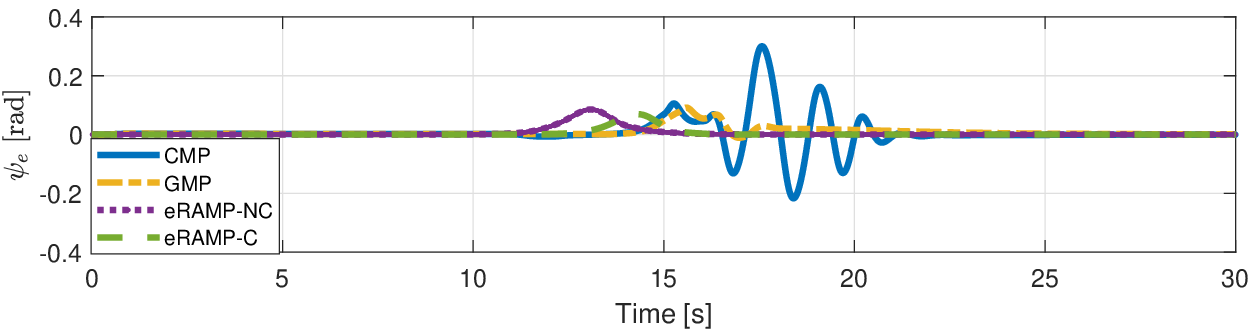}
        \caption{Heading angles of the ego vehicle.}
        \label{psi}
    \end{minipage}
\end{figure*}
\noindent The initial response of Definition \ref{def5} indicates that the ego vehicle can join the adjacent platoon from various locations if the front obstacle vehicle remains in the current lane for braking avoidance. Additionally, it should be noted that the spacing policy in ACC cannot be determined solely by safety, as the vehicle platoon should allow for closer headway between vehicles, thereby increasing highway capacity. Therefore, we apply the constant time gap (CTG) policy with a measured safe minimum distance for the vehicle platoon, taking into account the communication delay \cite{Hu2021-ip, Lee2021-ed}. If the front obstacle vehicle changes lanes for steering avoidance, Definition \ref{def5}.2.A indicates that the ego vehicle should follow RSS's first and second rules when joining the platoon from the front. Additionally, Definition \ref{def5}.2.B suggests that joining the platoon from the front and middle would be too dangerous due to insufficient space when the ego vehicle drives parallel to the platoon, making it safer to join from behind. Finally, the proposed motion planner can output the desired trajectory for the ego vehicle and the cooperative speed for adjacent connected AVs.

\section{Simulation Results and Analysis}\label{AA}

This section describes the simulation settings and results from the analysis. The simulation study was conducted on a desktop PC with Intel(R) Core(TM) i9-10980HK CPU@2.40GHz and RAM 32GB. Besides, four different motion planners were presented for comparative studies: (\romannumeral1) Conventional motion planner with constant speed, denoted as CMP \cite{Li2022-sz}; (\romannumeral2) Gaussian-based motion planner with speed planning, denoted as GMP \cite{Ji2023-gj}; (\romannumeral3) eRSS-RAMP method in non-cooperative, denoted as eRAMP-NC;(\romannumeral4) eRSS-RAMP method in cooperative, denoted as eRAMP-C.
\begin{figure*}[t]
\centering
    \begin{minipage}[t]{0.48\textwidth}
        \includegraphics[width=\hsize]{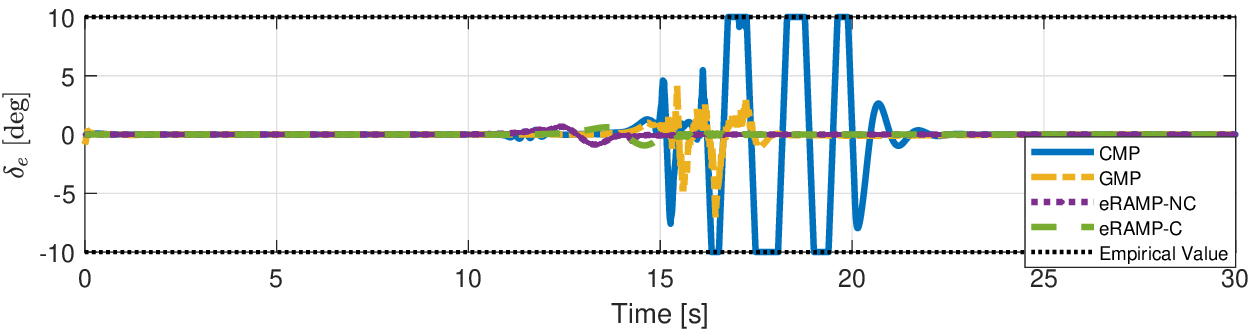}
        \caption{Front tire steering angles of the ego vehicle.}
        \label{delta}
    \end{minipage}\hspace{0.15in}
    \begin{minipage}[t]{0.48\textwidth}
        \includegraphics[width=\hsize]{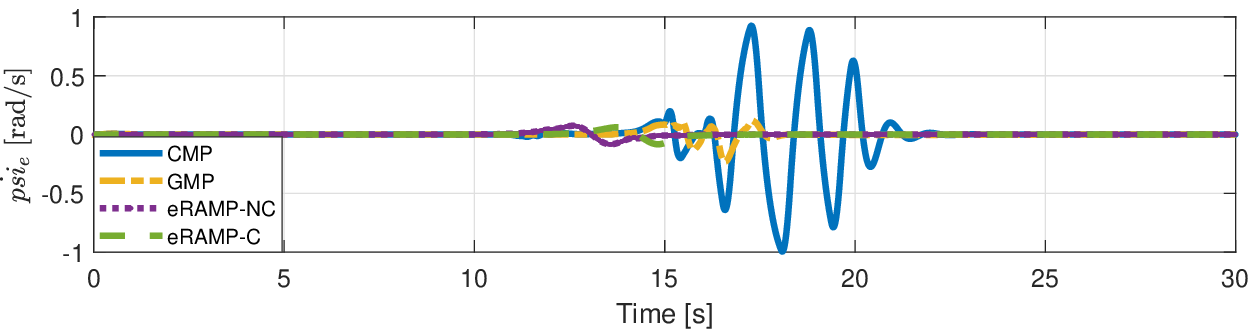}
        \caption{Heading rate of the ego vehicle.}
        \label{psi_rate}
    \end{minipage}
\end{figure*}
\begin{figure*}[t]
\centering
    \begin{minipage}[t]{0.48\textwidth}
        \includegraphics[width=\hsize]{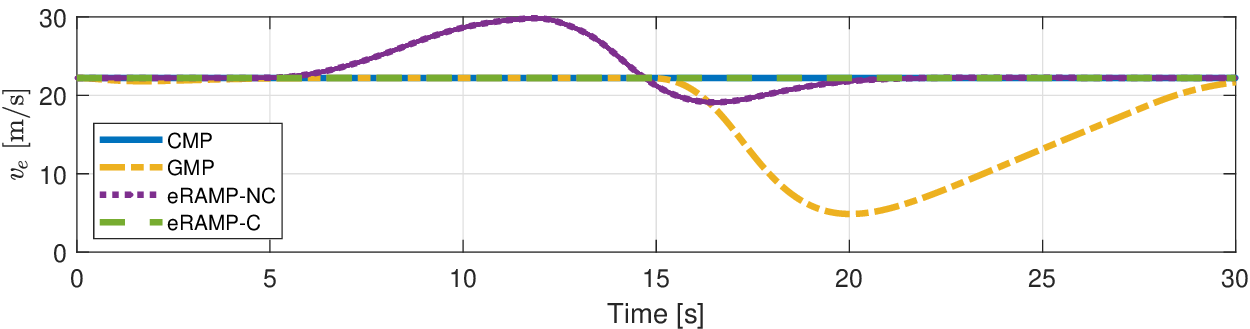}
        \caption{Longitudinal speeds of the ego vehicle.}
        \label{Vx_ego}
    \end{minipage}\hspace{0.15in}
    \begin{minipage}[t]{0.48\textwidth}
        \includegraphics[width=\hsize]{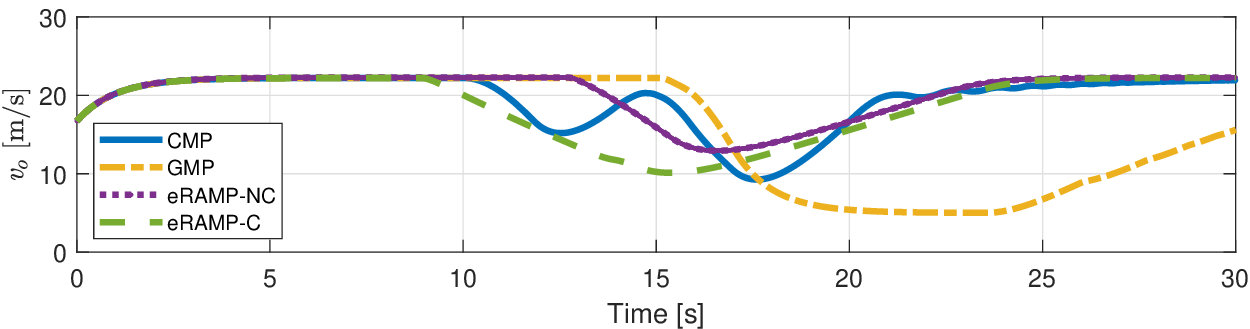}
        \caption{Longitudinal speeds of the obstacle vehicle.}
        \label{Vx_obs}
    \end{minipage}
\end{figure*}
\begin{table}[t]
    \centering
    \caption{Performance Evaluations}
    \setlength{\tabcolsep}{7.6pt}
    \begin{tabular}{cccccc}
    \hline
    Lane-Merge Planners                        & ML  & MT  & AMS  & C  & PO \\ \hline
    CMP                 & 300.49           & 13.52             & 22.22                               & 2.2e-2    &  \ding{51}    \\
    GMP               & 160.23           & 12.81              & 12.50                          & 0.8e-2      &  \ding{51}     \\ \hline
    eRAMP-NC & 128.25            & 4.80              & \textbf{26.72}             & \textbf{0.2e-2}        & \ding{55} \\ 
    eRAMP-C     & \textbf{75.24}   & \textbf{3.39}     & 22.22                      & 0.4e-2      & \ding{55}    \\ \hline
    \end{tabular}
    \begin{tablenotes}
        \footnotesize
        \item[*] ML: Merge Length (m) \quad MT: Merge Time (s) \quad AMS: Average Merge Speed (m/s) \quad C: Curvature (1/m) \quad PO: Path Oscillation
    \end{tablenotes}
    \label{indice}
\end{table}

\subsection{Lane-merge Scenarios}

For lane-merge scenarios, we set up an adjacent vehicle driving on the main lane with an initial speed of 16 m/s with a target speed of 22 m/s.We positioned the ego vehicle on the auxiliary road, relatively close to the obstacle vehicle, to simulate a more urgent lane-merge scenario under interaction uncertainty, as depicted in Fig. \ref{path}. We have evaluated those motion planners in the following performance indices, including merge length, merge time, average merge speed, curvature, and path oscillation, as shown in Table. \ref{indice}.

The eRAMP-C planner demonstrates the shortest merge length and merge time at 75.24 m and 3.39 s, respectively, indicating more efficient space usage during merging compared to other planners, particularly CMP, which exhibits the longest merge length and merge time at 300.49 m and 13.52 s. The highest average merge speed is achieved by eRAMP-NC at 26.72 m/s, while GMP has the lowest at 12.50 m/s, indicating that the eRAMP-NC planner facilitates a faster merging process, beneficial in high-speed scenarios. The eRAMP-C planner completed the merging maneuver at a constant speed due to the early deceleration of the adjacent vehicle. The planners exhibit various curvatures, with eRAMP-NC having the lowest at 0.2e-2 1/m, indicating smoother path trajectories. Conversely, GMP has a higher curvature, implying sharper turns during merging. Both eRAMP-NC and eRAMP-C show no path oscillation, while CPF and GPF do, suggesting that the eRAMP planners offer a more stable and consistent merging path. The performance indices from the table can be partially observed in the paths, as depicted in Fig. \ref{path}. We observe that the paths of the CMP planner and the obstacle vehicle exhibit significant oscillations in the merging zone (light gray area). The GMP planner achieved a smoother path; however, the path of the obstacle vehicle still oscillates. 

The motion states are listed from Fig. \ref{beta} to Fig. \ref{Vx_obs}. The CMP planner shows significant oscillations in sideslip angles after 15 s, peaking at around 0.15 rad, indicating instability. In contrast, the GMP, eRAMP-NC, and eRAMP-C planners maintain near-zero sideslip, reflecting better stability. The heading angle for CMP also exhibits pronounced oscillations post-15 s, while the other planners, especially eRAMP-C, show minimal deviations, suggesting smoother heading control. In Fig. \ref{delta}, The CMP planner again shows large oscillations in front tire steering angles, with peaks reaching the empirical values (10 deg). While the GMP planner has a shorter period of oscillation from 15 s to 17.5 s, with a peak at -7.0 deg. The eRAMP planners maintain relatively stable steering angles within $\pm$1 deg, indicating more controlled maneuvers. Similarly, the CMP planner’s heading rates fluctuate significantly after 15 seconds, contrasting with the more stable rates shown by the eRAMP-C planner. In Fig. \ref{Vx_ego}, the ego vehicle's speed under the eRAMP-NC planner reaches approximately 30 m/s, showing a steady increase. In contrast, the GMP planner's speed decreases sharply to below 10 m/s by 15 s before gradually increasing again. The CMP and eRAMP-C planners maintain relatively stable speeds near 20 m/s throughout. In Fig. \ref{Vx_obs}, the obstacle vehicle's speed varies more significantly under the GMP planner, dropping below 10 m/s before gradually recovering. The CMP planner also hovers around 20 m/s with conspicuous fluctuations. In contrast, the eRAMP-NC and eRAMP-C planners maintain steadier speeds, indicating better control over vehicle dynamics during the merging process.
% Two-lane 
\begin{figure*}[t]
    \centering
    \includegraphics[width=0.77\hsize]{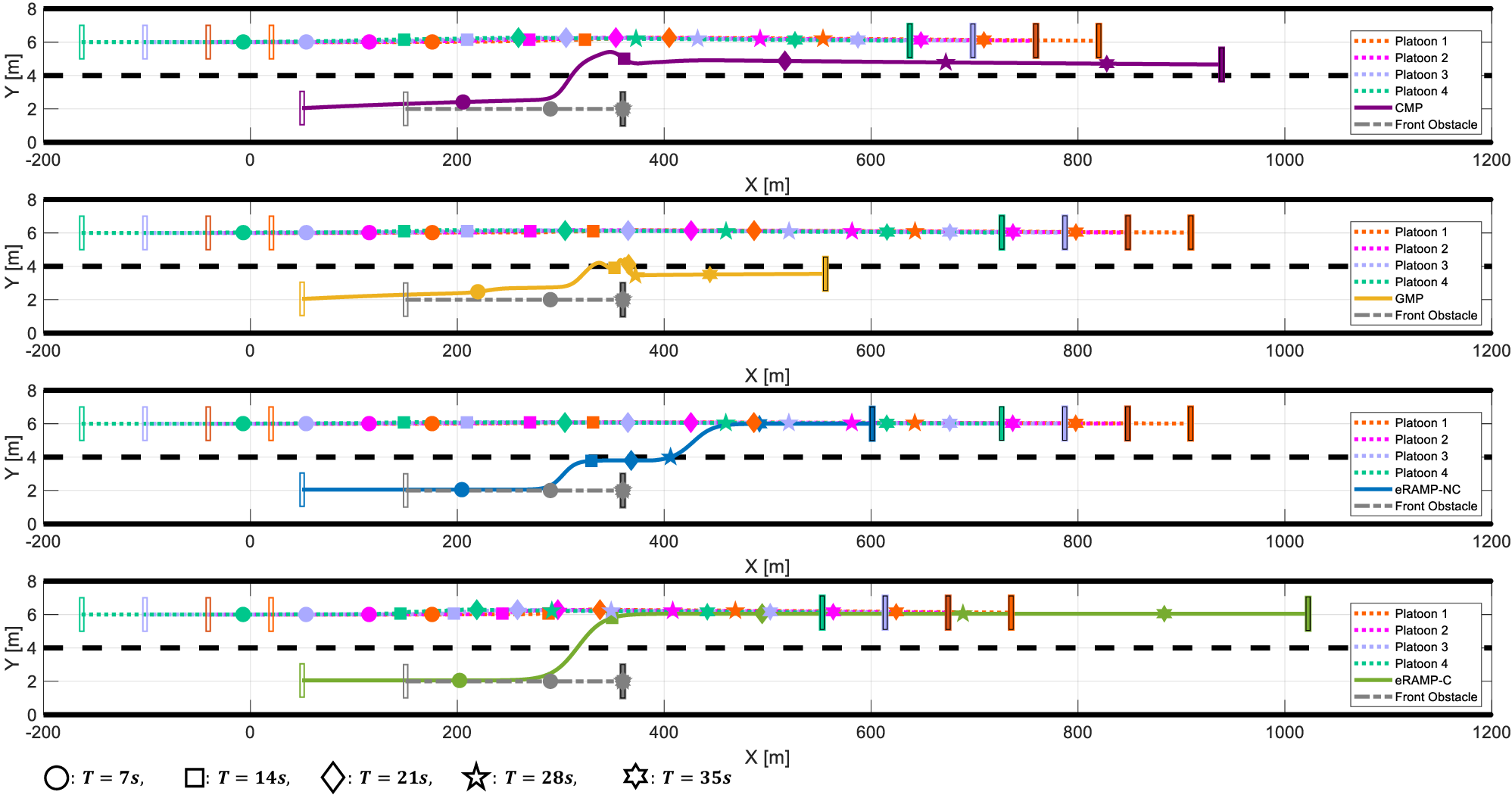}
    \caption{Obstacle avoidance trajectories based on four planners: CMP, GMP, eRAMP-NC, and eRAMP-C.}
    \label{path_two}
\end{figure*}
\begin{figure*}[t]
\centering
    \begin{minipage}[t]{0.48\textwidth}
        \includegraphics[width=\hsize]{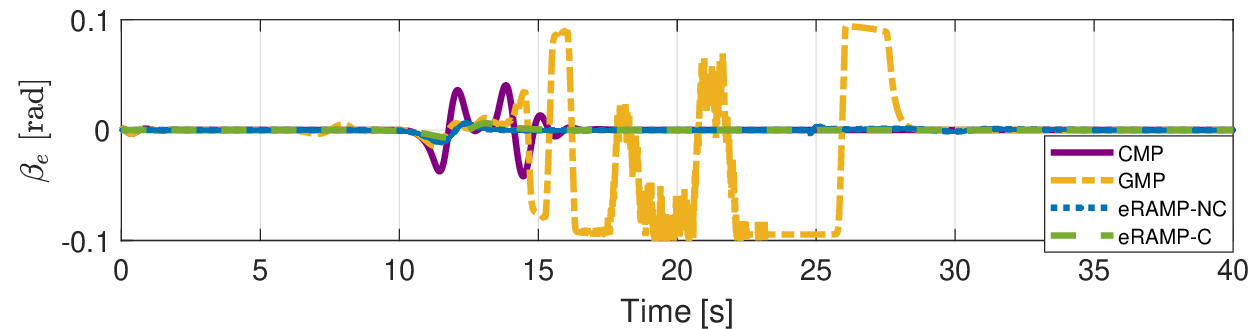}
        \caption{Sideslip angles of the ego vehicle.}
        \label{beta_two}
    \end{minipage}\hspace{0.15in}
    \begin{minipage}[t]{0.48\textwidth}
        \includegraphics[width=\hsize]{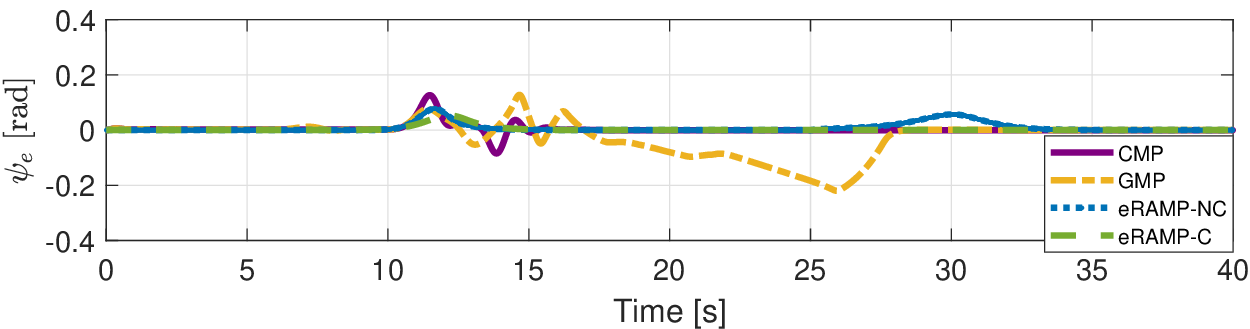}
        \caption{Heading angles of the ego vehicle.}
        \label{psi_two}
    \end{minipage}
\end{figure*}

\subsection{Emergency Collision Avoidance Scenarios}

\begin{figure*}[t]
\centering
    \begin{minipage}[t]{0.48\textwidth}
        \includegraphics[width=\hsize]{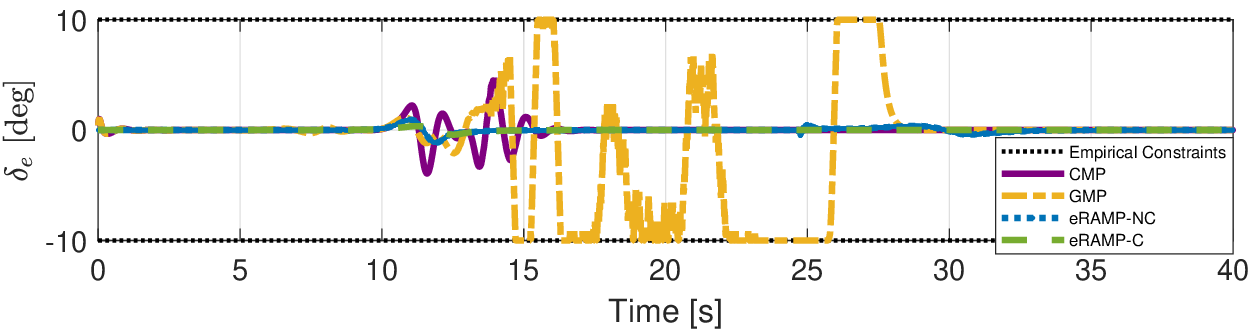}
        \caption{Front tire steering angles of the ego vehicle.}
        \label{tire_two}
    \end{minipage}\hspace{0.15in}
    \begin{minipage}[t]{0.48\textwidth}
        \includegraphics[width=\hsize]{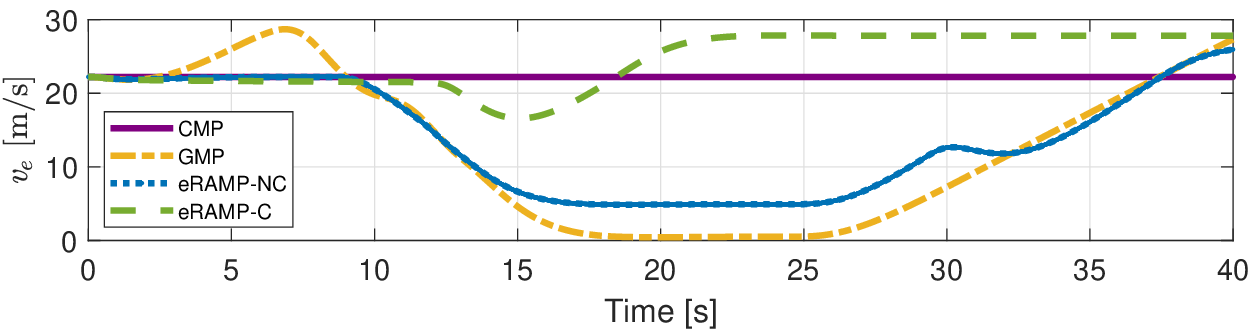}
        \caption{Longitudinal speeds of the ego vehicle.}
        \label{v_ego_two}
    \end{minipage}
\end{figure*}

\begin{figure*}[t]
\centering
    \begin{minipage}[t]{0.48\textwidth}
        \includegraphics[width=\hsize]{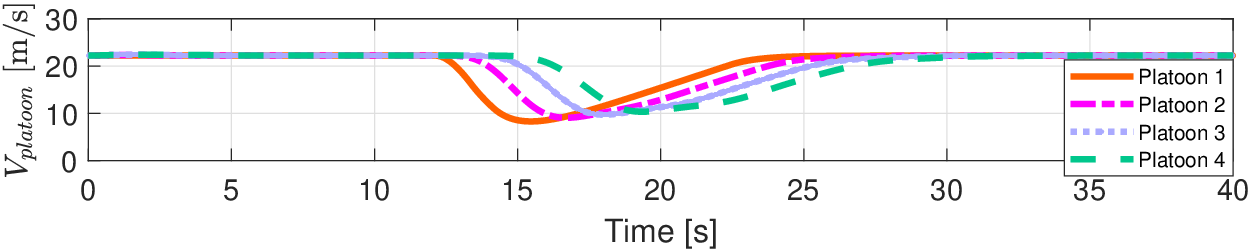}
        \caption{Longitudinal speeds of the vehicle platoon based on the CMP planner.}
        \label{v_platoon_ori}
    \end{minipage}\hspace{0.15in}
    \begin{minipage}[t]{0.48\textwidth}
        \includegraphics[width=\hsize]{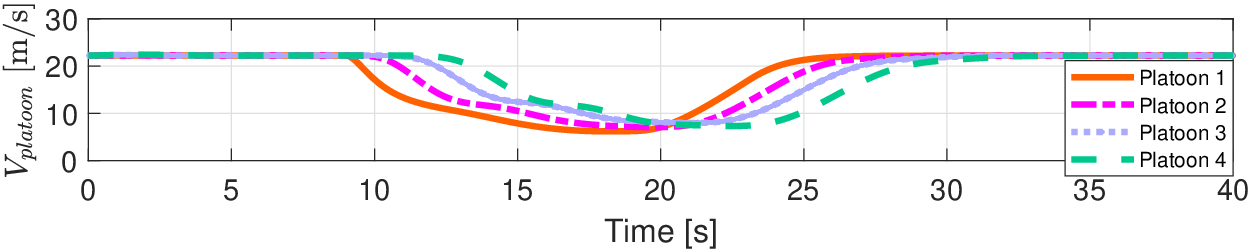}
        \caption{Longitudinal speeds of the vehicle platoon based on eRAMP-C.}
        \label{v_platoon_coop}
    \end{minipage}
\end{figure*}

To simulate an emergency obstacle avoidance driving scenario, we placed an obstacle vehicle 100 m ahead of the ego vehicle in the same lane and positioned a vehicle platoon in the adjacent lane approximately 30 m behind the ego vehicle longitudinally. The ego vehicle and the platoon started at an initial speed of 22.22 m/s, while the obstacle vehicle started at an initial speed of 20 m/s and would suddenly perform an emergency brake after traveling a certain distance. The paths of the four motion planners are shown in Fig. \ref{path_two} where we can observe that the CMP and GMP have obvious oscillations when processing the lane changing. Meanwhile, the trajectories of the platoon in the CMP planner are also influenced by the ego vehicle, resulting in a certain degree of lateral displacement; and the longitudinal speeds of the platoon decrease suddenly from 22.22 m/s to 8.31 m/s, plummeting by 13.91 m/s in 3 seconds, as depicted in Fig. \ref{v_platoon_ori}. While with the GMP planner, the longitudinal speeds of the platoon remain constant since the ego vehicle decelerates to an immediate stop, as shown (orange-yellow dashed line) in Fig. \ref{v_ego_two}. In contrast, the eRAMP planners complete smooth collision-free paths, where the eRAMP-NC planner decelerates to 4.92 m/s (Fig. \ref{v_ego_two}) and travels slowly in the middle of the road before accelerating to complete the lane-change process, as shown in Fig. \ref{path_two}. While the eRAMP-C planner has a minor deceleration from 21.47 m/s to 16.59 m/s, and then accelerates to 27.85 m/s after the lane change, as depicted in Fig. \ref{v_ego_two}. Additionally, the longitudinal speeds of the platoon shown in Fig. \ref{v_platoon_coop} indicate that the leading vehicle decelerates earlier, about 9.05 s, and at a slower rate.

The motion states are illustrated from Fig. \ref{beta_two} to Fig. \ref{tire_two}. The sideslip angle of the CMP planner exhibits a short period of fluctuation from 10.6 s to 15.8 s, peaking at 0.04 rad, while that of the GMP planner shows a longer period of fluctuation, reaching the empirical bound around $\pm$0.1 rad. In contrast, the eRAMP planners demonstrate a smaller variation in the sideslip angle within $\pm$0.012 rad, achieving more stable vehicle control. Consequently, the CMP planner shows a short-term fluctuation in the heading angle, peaking at 0.13 rad at about 11.5 s, as shown in Fig. \ref{psi_two}. While the GMP planner oscillates from 10.6 s to 17.4 s, then heads toward the right (down to -0.22 rad) to avoid the upcoming platoon. In contrast, the heading angles of the eRAMP planners vary within the positive half-plane, with all variations within 0.08 rad. Similar variations can also be observed in the front tire steering angles. The GMP planner exhibits significant oscillations, with amplitudes reaching the constraint value of 10 deg. The CMP planner shows relatively smaller oscillations between 10 s and 15 s. In contrast, the eRAMP planners demonstrate no oscillations, achieving smooth tire steering operations.

\section{Conclusion and Discussion}

In conclusion, our study demonstrates that the rule-adherence motion planner based on extended responsibility-sensitive safety significantly improves collaborative planning and collision avoidance in scenarios with interaction uncertainty. The eRSS-RAMP integrates the Responsibility-Sensitive Safety framework with potential fields and considers vehicle-to-vehicle communication delays and obstacle motion status for risk modeling. The proposed method achieves faster and safer lane merging performance, with a 53.0\% shorter merging length and a 73.5\% decrease in merging time, and allows for more stable steering maneuvers in emergency collision avoidance, resulting in smoother paths for the ego vehicle and surrounding vehicles.

Practically, the proposed eRSS-RAMP method improves safety and efficiency in lane merging and emergency avoidance, achieving faster and safer maneuvers. However, there are limitations: the reliance on simulations means real-world testing is needed to validate these findings, and the study mainly focuses on merging and emergency scenarios, leaving other critical scenarios, such as oncoming traffic and emergency cut-ins, for future research.

Future research should focus on further refining the eRSS-RAMP to address additional driving scenarios, such as oncoming traffic and emergency cut-in situations, and enhance real-world applicability by implementing and testing the method on real vehicles in autonomous driving test sites.
\vspace{-0.5cm}
% \section*{Acknowledgment}

% These research results were obtained from the commissioned research by the National Institute of Information and Communications Technology (NICT), JAPAN.

% \begin{thebibliography}{00}
% \end{thebibliography}
\bibliographystyle{IEEEtran}
\bibliography{IEEE_T_ITS_Ref.bib}

\end{document}